%% file: Lost_in_Conversation_or_Lost_in_Translation_Diagnosing_Multi_Turn_Degradation_in_RAG.tex
\documentclass{article}
\usepackage{iclr2027_conference,times}
\input{math_commands.tex}

\usepackage{hyperref}
\usepackage{url}
\usepackage{enumitem}
\usepackage{booktabs}
\usepackage{graphicx}
\usepackage{tabularx}
\usepackage[T1]{fontenc}
\usepackage[most]{tcolorbox}
\usepackage{listings}
\usepackage{float}
\usepackage{xcolor}
\definecolor{citeblue}{HTML}{2A5DB0}
\hypersetup{
  colorlinks=true,
  citecolor=blue,
  linkcolor=blue,
  urlcolor=blue,
}
\usepackage{wrapfig}
\newcommand{\sh}[1]{\setlength{\fboxsep}{1.2pt}\fcolorbox{black!45}{black!6}{\strut\scriptsize #1}}
\usepackage{pifont}
\usepackage{colortbl}
\usepackage{fontawesome5}
\usepackage{microtype}
\usepackage{soul}

\tcbuselibrary{listings}

\newtcblisting{configbox}[1]{%
  colback=gray!12,
  colframe=black,
  coltitle=white,
  colbacktitle=black,
  fonttitle=\bfseries,
  title={#1},
  boxrule=0.8pt,
  arc=2pt,
  left=5pt, right=5pt, top=3pt, bottom=3pt,
  listing only,
  listing options={
    basicstyle=\ttfamily\scriptsize,
    breaklines=true,
    columns=fullflexible,
    keepspaces=true
  }
}

\usepackage{fancyvrb}
\usepackage{fvextra}
\definecolor{shardgreen}{HTML}{3d8b5f}
\definecolor{simblue}{HTML}{2c5f9e}
\definecolor{assistgold}{HTML}{a07708}
\definecolor{judgered}{HTML}{c0392b}
\definecolor{retpurple}{HTML}{7b4fa0}
\newtcolorbox{promptbox}[2][]{enhanced, breakable,
  colframe=#2, colback=#2!5, colbacktitle=#2,
  boxrule=0.9pt, arc=2pt, left=5pt, right=5pt, top=3pt, bottom=3pt,
  fonttitle=\bfseries, title={#1}}
\usepackage[scaled=0.85]{FiraMono}
\DeclareTextCommandDefault{\textasciigrave}{\char`\`}
\title{Lost in Conversation or Lost in Translation? Diagnosing Multi-Turn Degradation in RAG}

\author{Pranav Handa \& Ariful Azad \\
Texas A\&M University \\
\texttt{\{pranavhanda2003, ariful\}@tamu.edu}}

\iclrfinalcopy

\begin{document}
\maketitle
\lhead{Preprint}

\begin{abstract}

When conversing with large language models (LLMs), users often begin with a
simple question and build towards a multi-hop question through
follow-up turns. 
Retrieval-augmented generation (RAG) and its graph-based variant (GraphRAG) have become the dominant approaches for grounding LLM responses in external evidence, yet both are evaluated almost exclusively on single-turn, fully specified queries.
We systematically investigate this evaluation mismatch through a
large-scale simulation study.
Building on prior work on multi-turn LLM evaluation, we transform questions from multi-hop question answering (QA) benchmarks into underspecified conversations and evaluate ten LLM assistants with eight retrieval systems across 1.5 million simulated conversations. 
Our findings reveal that multi-turn interaction causes widespread performance degradation, incurring relative performance drops of up to 21\% and increasing unreliability by 47\%, making RAG systems simultaneously less accurate and less reliable. 
We identify two distinct failure modes behind this degradation. Systems are either lost in translation, where conversational rephrasing distorts the retrieval query, or lost in conversation, where retrieval succeeds but the LLM fails to synthesize evidence distributed across turns.

\end{abstract}

\section{Introduction}

Large language models (LLMs) are increasingly used as interactive conversational agents, where users iteratively refine queries and add new details across multi-turn exchanges~\citep{herlihy2024overcoming, zhao2024wildchat, zheng2024lmsys}. 
Although this reflects how people naturally interact, recent work shows that it can severely degrade LLM performance. 
\citet{laban2026llms} demonstrate that incrementally revealing a fully specified instruction across multiple turns leads to an average 39\% performance drop across six generation tasks, a phenomenon they term \emph{lost in conversation}. 
Do retrieval-augmented generation (RAG) systems, which ground answers in documents retrieved from an external corpus, suffer the same degradation?
Existing question-answering benchmarks cannot resolve this, since they evaluate models almost exclusively on single, fully specified questions~\citep{yang2018hotpotqa, ho2020constructing, trivedi2022musiquemultihopquestionssinglehop, trivedi2023interleaving, gutierrez2024hipporag}. 
As RAG usage becomes widespread, we therefore ask: \emph{how much answer quality does a RAG system lose when the same information is revealed gradually through conversation rather than provided upfront?}

The question matters especially for RAG, because conversation changes not only what the LLM receives but also what gets retrieved. 
Retrieval can be sensitive to how a query is phrased \citep{penha2022evaluating, ma2023query}, and irrelevant or distracting evidence can hurt downstream generation \citep{shi2023large, yoran2024making,cuconasu2024power,liu2024lost}. 
In a conversational setting, where partial clues repeatedly become retrieval queries, these vulnerabilities can compound \citep{wu2022conqrr, mo2024chiq}. 
Recent benchmarks have begun to evaluate retrieval over evolving dialogues~\citep{anantha2021open,cheng2025coral,ali2026recor}, but they do not pair each conversation with a fully specified version of the same underlying question. 
As a result, current evaluations cannot quantify the performance degradation caused by conversational delivery.

We address this gap through a large-scale study built around multi-hop question answering. Multi-hop questions provide a natural testbed because their supporting facts can be separated and revealed progressively \citep{min2019multi, press2023measuring} while the original question remains available as a matched single-turn reference. We transform 750 questions from five multi-hop datasets into conversations and evaluate ten LLMs across eight retrieval settings, comprising a closed-book baseline and seven prominent retrieval methods spanning lexical, dense, hierarchical, and graph-based retrieval \citep{robertson2009probabilistic, sarthi2024raptor, gutierrez2024hipporag, gutierrez2025rag, ma2025think, guo2025lightrag}. In total, this amounts to 1.5 million simulated conversations. 

To localize where performance degrades, we evaluate three matched conditions from \citet{laban2026llms}: \textsc{FULL} (the original fully specified question), \textsc{CONCAT} (all conversational clues concatenated into a single turn), and \textsc{SHARDED} (clues revealed incrementally across turns). 
The performance drop from \textsc{FULL} to \textsc{CONCAT} isolates sensitivity to question reformulation, while the additional drop from \textsc{CONCAT} to \textsc{SHARDED} isolates the cost of multi-turn delivery. 
We further compare the retrieved evidence with LLM answer quality to distinguish failures in retrieval from LLM's failure to use that evidence. Finally, to account for the stochastic nature of LLM responses, we measure reliability across repeated simulations.

Our results show that the multi-turn gap is widespread. When the same question is delivered through conversation, answer quality drops by 11.7\% on average, with relative losses reaching 21\% in the hardest-hit settings. 
More importantly, our decomposition reveals two distinct sources of this degradation. For some systems, much of the loss occurs before any multi-turn interaction, when the original question is reformulated into conversational clues. We refer to this failure mode as \emph{lost in translation}. The effect is most pronounced for retrievers that operate directly on the query text or its embeddings. For other systems, the supporting evidence is successfully recovered across turns, but the LLM still fails to integrate it into a correct answer. We call this failure mode \emph{lost in conversation}. 

Beyond these failure modes, we also observe that strong single-turn performance does not guarantee conversational robustness. 
Our best-performing retriever degrades the most, and pairing it with the strongest LLM makes no difference. Consistent with \citet{laban2026llms}, we find that conversation also makes system behavior substantially less reliable. Run-to-run unreliability increases by 47\%, even as performance in the strongest runs remains relatively stable.
These results show that standard single-turn evaluations can miss failures that arise both during query reformulation and after the relevant evidence has been retrieved. 

Interestingly, we find that this degradation can be partly undone. Simple agent-style interventions that repeat information from earlier turns nearly erase the gap for a subset of retrievers that share a common preprocessing step. This full gain does not appear in the LLM-only setting of \citet{laban2026llms}, and is thus specific to these retrieval systems.

The main contributions of this paper are:
\begin{itemize}[leftmargin=20pt]

    \item \textbf{A multi-turn benchmark for RAG.} We convert 750 questions from five multi-hop QA benchmarks into manually reviewed conversations, each paired with its original fully specified question. We use it to evaluate ten LLMs with seven retrieval methods and a closed-book baseline.

    \item \textbf{Empirical evaluation of RAG under multi-turn interaction.} Across 1.5 million simulated conversations, we observe an average answer quality drop of 11.7\%. We trace this degradation to two distinct failure modes: \emph{losing the question in translation} during initial reformulation and \emph{losing the evidence in conversation} across multi-turn delivery.
    
    \item \textbf{A path toward more robust multi-turn RAG.}  We show that a
    lightweight, training-free intervention recovers much of the lost
    performance, but only for retrievers with a particular property that we
    identify, giving RAG system builders a concrete guideline for choosing
    designs that stay robust across conversations.

\end{itemize}

\section{Background and Related Work}

Retrieval-augmented generation has advanced rapidly and become a standard way to equip language models for specialized knowledge-intensive tasks~\citep{lewis2020retrieval,gao2023retrieval}. Recent work continues to improve how evidence is retrieved and used for multi-hop reasoning~\citep{wang2026chain, guan2026deeprag,gutierrez2025rag, luo2026gfm,ma2025think}.

In practice, these systems are increasingly used inside conversational assistants such as ChatGPT and Claude, where a user rarely states their full need at once, and instead arrives at it through clarifications and follow-ups over several turns. This has prompted a growing line of work on conversational RAG, with recent benchmarks evaluating retrieval and generation over multi-turn interactions~\citep{katsis2025mtrag,cheng2025coral, ali2026recor, rosenthal2026mtrag}. Related work has also studied conversational reformulation, where earlier turns are used to resolve and rewrite context-dependent queries for retrieval~\citep{wu2022conqrr, mo2024chiq}. Together, this work moves RAG evaluation closer to real conversational use. However, these benchmarks evaluate systems only in the multi-turn setting, so it is hard to tell how much of the error comes from the conversation itself versus the difficulty of the underlying questions.

A separate line of work addresses this directly by asking a more targeted question. How much performance is lost when the same task is spread across multiple turns rather than presented all at once? Comparing the two settings lets us separate the effect of conversation from the difficulty of the task. \citet{kwan2024mt} construct single-turn counterparts for multi-turn interaction and show that strong single-turn performance does not necessarily translate to multi-turn settings. More recently, \citet{laban2026llms} find an average 39\% performance drop in a similar setting, across basic generation tasks such as code generation and multi-document summarization. They further show that this drop comes primarily from increased unreliability of LLMs rather than a loss of aptitude. Both studies, however, look at the language model on its own, without retrieval. We extend their setup to retrieval-augmented QA, where splitting a question across turns affects not just how a model reasons but also what evidence gets retrieved at each turn. This lets us measure how much each retrieval method loses in conversation and understand why some hold up better than others.

\section{Constructing Multi-Turn Evaluations from Multi-Hop QA}
\label{sec:benchmark}
Each instance of our benchmark starts from a fully specified multi-hop
question $q$, which we decompose into an ordered sequence of \emph{shards}
$s_1, \ldots, s_k$. The first shard states what the user is looking
for, and each subsequent shard introduces one additional clue from $q$.
Together, these shards preserve the information in the original question, but
distribute it across multiple turns (Section~\ref{sec:sharding}).

Each simulated conversation involves three LLM components: a \emph{user
simulator}, the \emph{LLM assistant} under evaluation, and a \emph{response
classifier}, which we describe in full in Appendix~\ref{app:user_sim}. The user simulator holds the full shard set and reveals it one
shard at a time. It opens with $s_1$, and on each later turn it rephrases the
next unrevealed shard into the user message $u_t$, conditioned on the conversation so far. Before the assistant responds at turn $t$, a retriever
$R$ queries a fixed corpus or graph $\mathcal{D}$. Conversational retrieval methods commonly rewrite context-dependent turns into standalone queries \citep{wu2022conqrr, mo2023convgqr}. Rather than introducing
a learned rewriter, we consider two transparent retrieval policies $P \in \{\textsc{current},\textsc{history}\}$, which determine the
query passed to the retriever: 
\[
x_t(P)=
\begin{cases}
u_t & P=\textsc{current},\\[2pt]
u_1 \oplus \cdots \oplus u_t & P=\textsc{history}.
\end{cases}
\]
where $\oplus$ denotes concatenation of the user turns.
The retriever then returns the evidence $r_t = R(x_t(P);\,\mathcal{D})$ for turn $t$, and the
LLM responds using the conversation history and this evidence:
\[
a_t = \text{LLM}(H_t,r_t),
\qquad
H_t=(u_1,a_1,\ldots,a_{t-1},u_t).
\]

After each LLM assistant turn, the response classifier identifies committed
answer attempts, which are parsed and scored against the gold answer. The
conversation stops when the assistant produces an exactly correct answer or
after it has responded to the final shard. If an answer is incorrect and
unrevealed shards remain, the simulator provides one additional clue on the
next turn.

\begin{figure}[ht]
\centering
\includegraphics[width=\linewidth]{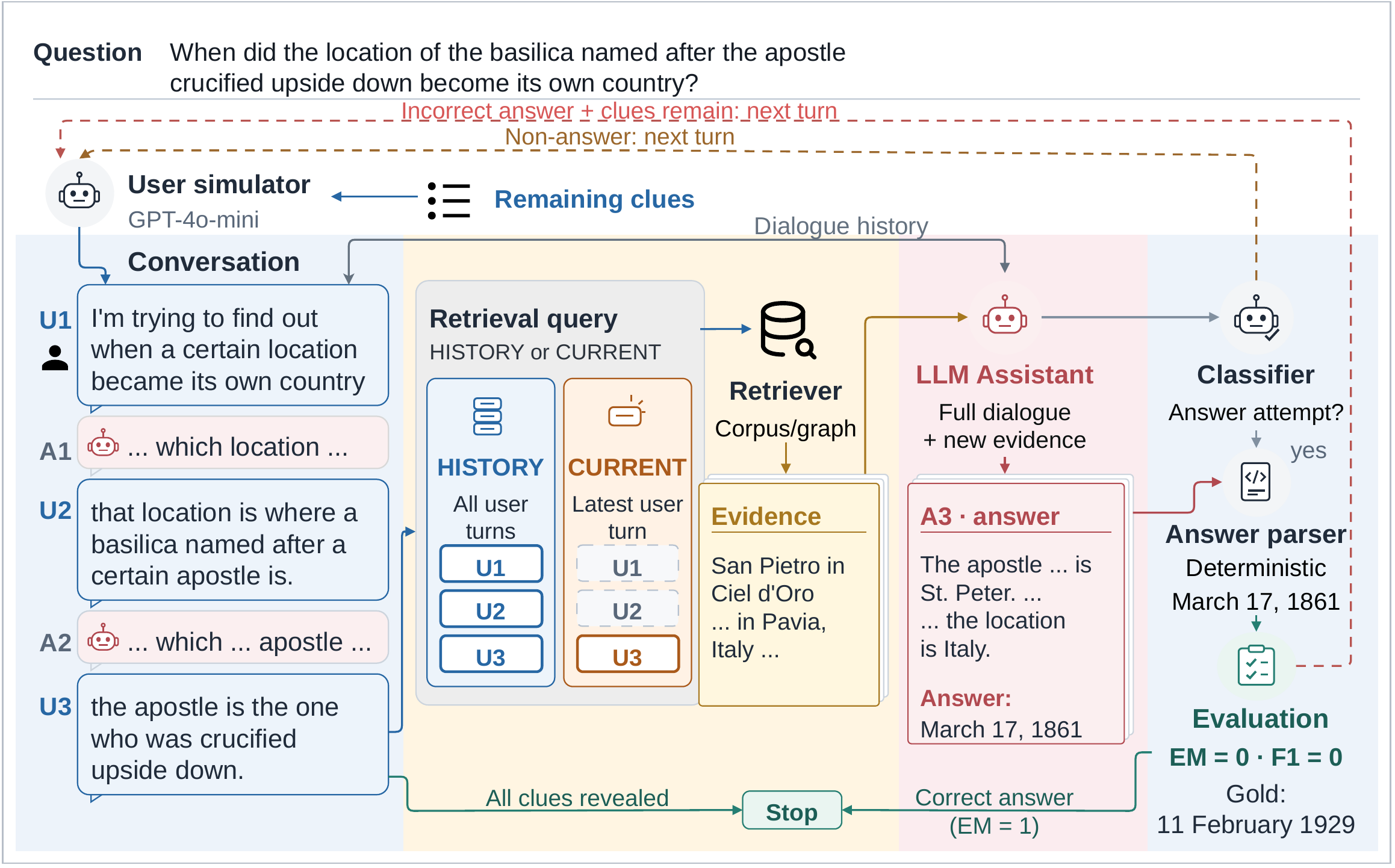}
\caption{
Overview of a multi-turn retrieval simulation, shown at turn~3. The user simulator has just revealed the next clue~($u_3$). The retriever is then queried with either all user turns so far ($u_1\!+\!u_2\!+\!u_3$) or only the latest~($u_3$), depending on the retrieval policy. The LLM then answers from this retrieved evidence together with the conversation history, and the response classifier flags its output as an answer attempt. The text after the \texttt{Answer:} marker is scored against the gold answer. Here the predicted date is wrong, so the assistant receives a score of 0 F1.
}
\label{fig:simulation_overview}
\vspace{-4ex}
\end{figure}

Figure~\ref{fig:simulation_overview} illustrates our simulation design, which adapts the multi-turn framework of \citet{laban2026llms} (hereafter the \emph{original benchmark}).
We keep its user simulation, response classification,
and insert a retrieval step at every turn. The user simulator and response
classifier are fixed GPT-4o-mini modules, while the LLM assistant and
retriever are the systems we evaluate.

\subsection{Datasets}
We draw on three multi-hop question answering datasets widely used in
retrieval evaluation: HotpotQA (HP)~\citep{yang2018hotpotqa},
2WikiMultiHopQA (2W)~\citep{ho2020constructing}, and MuSiQue
(MS)~\citep{trivedi2022musiquemultihopquestionssinglehop}.
In addition, we separate MuSiQue by reasoning depth into 2-hop, 3-hop, and 4-hop subsets (MS-2, MS-3, MS-4). Deeper questions require longer chains of evidence and also produce longer sharded conversations. This lets us observe multi-turn degradation across conversations of varying length and reasoning depth.

We evaluate 150 questions from each of these five subsets, for 750 reviewed questions
in total. For retrieval, HotpotQA and 2Wiki each use a fixed corpus of 1,000 questions,
while MuSiQue uses a single shared corpus of 1,000 questions spanning its
three hops. This 1,000-question corpus size matches the setting on which these
methods are commonly evaluated~\citep{gutierrez2024hipporag,gutierrez2025rag,luo2026gfm,santhana2026browsenet}.
The reviewed questions are included in these pools, while the rest contribute
candidate passages but are not scored.

\subsection{From a Full Question to a Conversation}
\label{sec:sharding}
Following the original benchmark's segment, rephrase, verify, and inspect
process, a GPT-4o-mini module first \emph{segments} each multi-hop question into its
individual units of information and then \emph{rephrases} each unit as a
conversational user turn. Both steps follow a set of rules we define to
ensure a fair comparison. Each resulting shard set is automatically
\emph{verified} and then manually \emph{inspected} against these rules.
Appendix~\ref{app:benchmark} details this process and the rules every valid
shard set must satisfy, and Appendix~\ref{app:prompts} lists the exact
prompts.

While the original benchmark assumes order-insensitive shards, multi-hop questions form reasoning chains that require sequential context. 
We therefore make two adaptations. First, we order the shards according to the question's compositional structure rather than
presenting them in an arbitrary order, so that later shards can refer back
to entities introduced earlier using expressions such as ``that person''
or ``that film.'' 
Second, we tighten the sharding constraints after
observing that the sharding model could leak intermediate answers into the
shard text or turn a clue
into a direct subquestion (e.g.\ ``who voices Glenn Quagmire?'') instead
of presenting it as a piece of the user's description (e.g.\ ``it was
created by whoever does the voice of Glenn Quagmire''). 
Shards must therefore preserve original entities and relations without introducing unmentioned facts or explicitly decomposing the query into sub-problems.
The final 750 shard sets were manually reviewed under these criteria.
\subsection{Simulation Types} 
\begin{wraptable}{r}{0.52\linewidth}
\vspace{-3ex}
\centering\small
\setlength{\tabcolsep}{4pt}
\caption{Information-delivery conditions. Each box represents one user turn.}\vspace{2pt}
\label{tab:conditions}
\begin{tabular}{@{}lp{0.70\linewidth}@{}}
\toprule
Condition & Assistant Input \\
\midrule
\textsc{FULL}     & The original question, in one turn\hfill\sh{$q$} \\
\textsc{CONCAT}   & All shards as a list, in one turn\hfill\sh{$u_1\,u_2\,u_3$} \\
\textsc{SHARDED}  & One shard per turn\hfill\sh{$u_1$}\hspace{2pt}\sh{$u_2$}\hspace{2pt}\sh{$u_3$} \\
\textsc{RECAP}    & \textsc{SHARDED}, then one final turn restating every shard\hfill\sh{$u_1$}\hspace{2pt}\sh{$u_2$}\hspace{2pt}\sh{$u_3$}\hspace{2pt}\sh{$u_1\,u_2\,u_3$} \\
\textsc{SNOWBALL} & \textsc{SHARDED}, with each turn repeating all earlier shards\hfill\sh{$u_1$}\hspace{2pt}\sh{$u_1\,u_2$}\hspace{2pt}\sh{$u_1\,u_2\,u_3$} \\
\bottomrule
\end{tabular}
\vspace{-2ex}
\end{wraptable}
Table~\ref{tab:conditions} describes 
five information-delivery conditions from the original benchmark, each of which presents the same underlying information in a different conversational form.

\textsc{FULL} against \textsc{SHARDED} is our central comparison. 
\textsc{CONCAT} lets us separate changes caused by shard formulation from those caused by distributing the same information across multiple conversational turns. Following \citet{laban2026llms}, we use closed-book \textsc{CONCAT} as an information-preservation
check, requiring
$\overline{\mathrm{F1}}_{\textsc{CONCAT}}
\geq 0.8\,\overline{\mathrm{F1}}_{\textsc{FULL}}$.

\section{Experimental Setup}
\label{sec:setup}

\subsection{Retrieval Methods}
Our comparison consists of eight retrieval settings over the shared corpus: a closed-book configuration (NONE), two passage retrievers (BM25 and dense vanilla RAG), one hierarchical retriever (RAPTOR), and
four graph retrievers (HippoRAG, HippoRAG2, ToG-2, LightRAG). All retrieval methods use bge-large-en-v1.5 where applicable, while retriever-specific index structures,
query-grounding procedures, and evidence units are preserved. Most passage-oriented systems
use top-five retrieval. Hierarchical and graph systems retain their respective retrieval units
and output rules. 

For graph-based retrievers, we build a graph from the same passage corpus using
an LLM-based OpenIE procedure. Following prior graph retrievers~\citep{gutierrez2024hipporag,gutierrez2025rag,luo2026gfm}, 
we extract entity mentions and relation-bearing triples from each passage. For methods 
that require graph seeds, the fixed query-NER LLM component extracts mentions from the visible 
query, and dense similarity links them to nodes in the corpus-local graph.

Methods originally designed for external knowledge graphs are adapted to this
shared graph so that all retrievers use identical source evidence. Implementation details are provided in Appendix~\ref{app:retrievers}.

\subsection{Language Models}
We evaluate ten LLMs from four model families: GPT-4o-mini and
5.6~Luna (OpenAI), Llama-3.1-8B-Instruct and Llama-3.3-70B-Instruct
(Meta), Gemma-4-31B-IT (Google), Qwen3-4B, Qwen3-8B, Qwen3-14B, Qwen3-32B, and Qwen3.6-27B (Alibaba). The selection spans
4B to 70B+ parameters across both API and open-weight models. 
Every model follows a shared output format in which candidate answers appear after a designated \texttt{Answer:} marker. 
Exact model versions and serving configurations are listed in Appendix~\ref{app:model_access}. Evaluating API models across all combinations of datasets, retrievers, query policies, and simulation types would be prohibitively expensive. We therefore use a prespecified set of API models and rely on open-weight models for broader model coverage.

\subsection{Evaluation Metrics}
\label{sec:metrics}

LLM generation is stochastic, and otherwise identical runs can produce different answers.
We simulate each question five times, which lets us account for the run-to-run variation while keeping the substantially larger model-by-retriever evaluation tractable. 

Answers are scored using normalized exact match (EM) and token-level
F1, following the evaluation used in each source
dataset~\citep{yang2018hotpotqa,trivedi2022musiquemultihopquestionssinglehop}. We retain the conversation-level BEST scoring convention of \citet{laban2026llms}. \textbf{BEST} is the maximum F1 over all answer attempts in a conversation. Conversations with no answer attempt receive zero. 
We also report paired 95\% confidence intervals for differences between conditions using bootstrap resampling over questions. 

\begin{table}[!th]
\centering
\caption{F1 ($\times 100$) under FULL and SHARDED, averaged over
ten LLMs. Red shading indicates
a loss and blue a gain relative to FULL; intensity reflects the absolute
difference in F1 points. Rows are ordered by mean FULL score.}
\input{retrieval_gap_laban.tex}
\label{tab:retrieval_gap}
\end{table}

\section{Results}
\label{sec:results}

\subsection{Multi-turn degradation is present without retrieval}
\label{sec:closed_book_gap}

We find that \textbf{multi-turn degradation is already present without retrieval}. Averaged over all ten LLMs and the five datasets, closed-book F1 falls from 0.230 under \textsc{FULL} to 0.190 under \textsc{SHARDED}, a relative decline of 17\%. For the largest model in each family, \textsc{CONCAT} retains 94.3\% of \textsc{FULL} performance, and all five datasets satisfy the information-preservation criterion. This extends the multi-turn sensitivity reported by
\citet{laban2026llms} from generation tasks to multi-hop QA. Appendix~\ref{app:closed_book} reports the full per-model and per-dataset breakdown.

\subsection{Retrieval does not eliminate multi-turn degradation}
\label{sec:not_eliminate}

\textbf{Multi-turn degradation persists with retrieval.}
Table~\ref{tab:retrieval_gap} shows that retrieval improves answer quality over the closed-book setting, yet every retrieval method has a positive mean \textsc{FULL}--\textsc{SHARDED} gap under both query policies. Averaged over ten LLMs, seven retrieval methods, and five datasets, we see a decline of
5.1 F1 points under \textsc{HISTORY} and 7.1 under \textsc{CURRENT}, corresponding to relative losses of 11.7\% and 16.3\%, respectively.

Notably, the strongest retrieval method suffers the largest absolute
degradation. HippoRAG2 attains the highest mean \textsc{FULL} score of
50.6, yet loses 6.8 F1 points under \textsc{HISTORY} and 9.9 under
\textsc{CURRENT}, more than any other method. This indicates that
\textbf{single-turn retrieval quality does not predict multi-turn
robustness.} The same pattern holds among methods with similar
single-turn performance. Dense and ToG-2 have mean \textsc{FULL}
scores of 45.4 and 44.4, yet lose 6.0 and 3.7 points under
\textsc{HISTORY}. In short, \textbf{each retrieval method degrades differently}, and single-turn evaluation conceals these differences,
underscoring the need to evaluate retrieval systems in multi-turn
settings.

\begin{figure}[th]
\centering
\includegraphics[width=\linewidth]{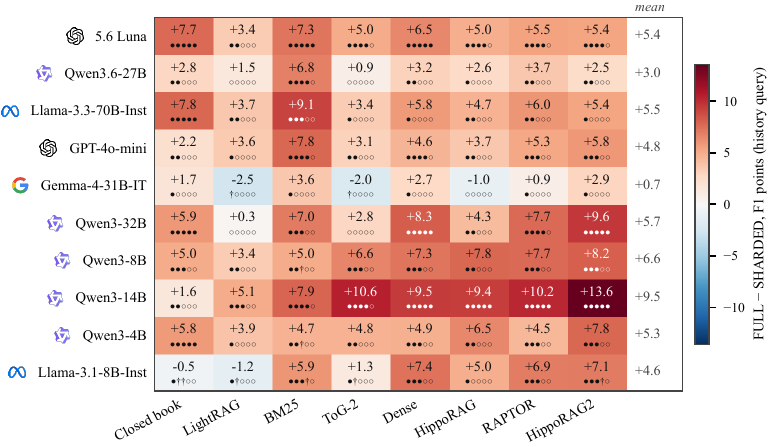}
\caption{\textsc{FULL} vs.\ \textsc{SHARDED} F1 (points) under history queries.
Each cell averages over the five datasets. Rows are ordered by mean
\textsc{FULL} score across the seven retrieval methods.
The five symbols beneath each value summarize dataset-level paired
95\% bootstrap intervals for $\Delta=\textsc{FULL}-\textsc{SHARDED}$:
$\bullet$ entirely above zero, $\dagger$ entirely below zero, and
$\circ$ containing zero. Intervals excluding zero indicate a statistically
significant difference at the 5\% level.}
\label{fig:gap_heatmap}
\vspace{-10pt}
\end{figure}

\textbf{The highest-scoring models are not necessarily the most robust.}
Figure~\ref{fig:gap_heatmap} shows that 5.6 Luna, the highest-scoring model under both \textsc{FULL} and \textsc{SHARDED}, still loses 5.4 F1 points on average under \textsc{HISTORY}, with significant declines in 28 of its 35 retriever--dataset pairs. Gemma-4-31B provides a contrasting example: despite ranking fifth under \textsc{FULL}, it loses only 0.7 F1 points on average, although its run-to-run unreliability nearly doubles (Appendix~\ref{app:reliability}). In fact, a model that scores 5.1 points above Gemma under \textsc{FULL} (Llama-3.3-70B) ends at essentially the same level under the multi-turn setting. These comparisons suggest that \textbf{a highly capable LLM can lose its entire advantage in a multi-turn
setting.} Multi-turn robustness therefore has to be tested under simulated underspecified conversations, not inferred from single-turn scores.

\textbf{More conversational history is not always a better retrieval query.}
A natural question is how much of the retrieval query should come from
earlier turns. We test this by comparing \textsc{HISTORY}, which passes every
user turn to the retriever, with \textsc{CURRENT}, which passes only
the latest one. Our results show that using previous user turns (i.e., \textsc{HISTORY}) improves mean \textsc{SHARDED}
performance for six of the seven retrieval methods. 
BM25 is the exception where switching to \textsc{CURRENT} reduces its mean gap from 6.5 to 3.6 F1 points. Its MuSiQue-3hop result is the only case where \textsc{SHARDED} exceeds \textsc{FULL}. \textbf{This indicates that earlier turns generally provide useful retrieval context, although the benefit depends on how the retriever
processes its query.} 
This is consistent with work showing that conversational history often needs to be selected, denoised, or otherwise resolved before retrieval \citep{mo2024history,mo2024chiq}. Section~\ref{sec:retrieval_coverage} shows what the two queries retrieve.

Based on these results, we attribute the degradation to two
forces. First, formulation, or being \emph{lost in translation}: converting a
fully specified question into conversational shards changes what the
retriever finds, even when the semantic content is preserved. Second,
the \emph{lost-in-conversation} effect of \citet{laban2026llms}:
distributing the same shards across turns may independently affect
both retrieval and the assistant's ability to integrate the evidence.
We isolate these effects next.

\subsection{Decomposing the Gap: Formulation and Multi-Turn Delivery}
\label{sec:concat_decomposition}
The \textsc{FULL}--\textsc{SHARDED} gap combines two changes: the formulation of the question and its delivery across turns. We use \textsc{CONCAT} as an intermediate condition to write the total gap as 
\begin{equation}
\begin{aligned}
\Delta
&= \overline{P}_{\textsc{FULL}}
 - \overline{P}_{\textsc{SHARDED}} \\
&=
\underbrace{
\overline{P}_{\textsc{FULL}}
-\overline{P}_{\textsc{CONCAT}}
}_{\Delta_{\mathrm{form}}}
+
\underbrace{
\overline{P}_{\textsc{CONCAT}}
-\overline{P}_{\textsc{SHARDED}}
}_{\Delta_{\mathrm{turn}}}.
\end{aligned}
\label{eq:gap_decomposition}
\end{equation}

Here, $\overline{P}$ denotes mean F1. 
The formulation contrast (${\Delta_{\mathrm{form}}}$) changes both the retrieval query and the assistant's input. The multi-turn contrast (${\Delta_{\mathrm{turn}}}$) additionally distributes the same shards across turns and introduces repeated retrieval and answer opportunities.

\begin{table}[th]
\centering
\caption{
Decomposition of the \textsc{FULL}--\textsc{SHARDED} F1 gap ($\times 100$) under \textsc{HISTORY} queries. All cells average over ten LLMs. FULL, Mean, and Total columns also average over the five datasets.
}
\input{decomposition_laban.tex}
\label{tab:gap_decomposition}
\end{table}

\textbf{Formulation losses differ sharply across retrievers.}
Table~\ref{tab:gap_decomposition} shows a positive mean $\Delta_{\mathrm{form}}$ for every retriever. BM25 has the largest drop, at 11.7 F1 points, while Dense, RAPTOR, and HippoRAG2 lose
7.1--8.1 points. 
By contrast, HippoRAG, LightRAG, and ToG-2 lose only 1.4--2.8 points. 
Interestingly, the latter three methods first use an LLM to extract entities or keywords from the query before retrieval. Since \textsc{FULL} and
\textsc{CONCAT} preserve the underlying entities and relations, this preprocessing may make them less sensitive to conversational rephrasing, although the advantage does not necessarily carry over once the evidence is distributed across turns.

\textbf{Multi-turn delivery can partially reverse formulation losses.} The pattern changes once the same shards are distributed across turns. BM25 recovers 5.2 F1 points over \textsc{CONCAT}, while RAPTOR, Dense, and HippoRAG2 recover 2.2, 1.2, and 0.3 points, respectively. The three retrievers that were least affected by formulation do not show the same recovery: HippoRAG loses a further 3.4 points, while LightRAG and ToG-2 also decline slightly.

We interpret this pattern as the interaction of two opposing effects within $\Delta_{\mathrm{turn}}$. Repeated retrieval gives formulation-sensitive methods additional opportunities to recover evidence missed by the single \textsc{CONCAT} query, offsetting part of their initial loss.
At the same time, distributing the question across turns adds a cost of its own. Even without retrieval, \textsc{SHARDED} falls 1.5 points below \textsc{CONCAT}. Thus, retrievers with larger formulation losses recover through repeated retrieval, but this recovery remains incomplete, and every retriever ultimately exhibits a net \textsc{FULL}--\textsc{SHARDED} degradation.

So far, both effects are inferred from answer scores alone. Next, we test whether the retrieved passages show the same pattern, using per-query and cumulative Recall@5.

\subsection{Retrieval Coverage across Turns}
\label{sec:retrieval_coverage}

\textbf{Question formulation changes which evidence is retrieved.}
We observe in Table~\ref{tab:recall} that mean Recall@5 falls from 56.7 under \textsc{FULL} to 49.2 under
\textsc{CONCAT}, but the effect again varies substantially across
retrievers. For instance, BM25 drops from 45.6 to 18.6, whereas HippoRAG
increases from 61.7 to 65.1. 
As with the reformulation results, this shows that changing a question's phrasing substantially changes the evidence retrieved for the LLM, whereas methods that preprocess the query with an LLM are less affected.

\textbf{Successive retrieval can recover missed evidence.}
Under \textsc{HISTORY}, cumulative recall reaches 55.8, close to the 56.7 \textsc{FULL} baseline. BM25 again provides the clearest example as its recall falls to 18.6 under \textsc{CONCAT}, but reaches 35.3 when retrievals are accumulated across \textsc{SHARDED} turns. This supports the explanation above that later retrievals can recover supporting evidence missed by the single \textsc{CONCAT} query, although the amount of recovery differs across retrievers.

\begin{table}[th]
\centering\footnotesize\setlength{\tabcolsep}{4.8pt}
\caption{
Supporting-passage Recall@5 ($\times 100$), averaged over ten LLMs and five datasets. For \textsc{SHARDED}, per query is the mean over retrieval turns, final turn uses the last retrieval, and cumulative is the union of every turn's top-five passages. Underlined entries exceed the row's \textsc{FULL} recall. 
RAPTOR is omitted because its retrieval units consist of normal passages and their summary nodes.
}
\input{recall_main.tex}
\label{tab:recall}
\vspace{-8pt}
\end{table}

\textbf{Coverage across turns does not guarantee answer recovery.}
Moreover, when we compare cumulative recall with answer performance, higher
retrieval coverage does not consistently correspond to higher F1.
Under \textsc{CURRENT}, cumulative recall reaches 61.0 and exceeds
\textsc{FULL} for every retriever, yet \textsc{SHARDED} answer
performance remains lower. For instance, HippoRAG2 reaches 74.8 cumulative
recall, compared with 70.7 under \textsc{FULL}, while its F1 remains
9.9 points lower. \textbf{In other words, retrieval coverage alone is not sufficient in multi-turn RAG; 
the LLM must also be able to piece together evidence that becomes available across separate turns.} We test this directly through a paired study (Appendix~\ref{app:evidence_coverage}): when a \textsc{FULL} and \textsc{SHARDED} conversation both retrieve all of their supporting passages, \textsc{SHARDED} still scores 11.6 F1 points lower under \textsc{HISTORY}.

\textbf{The multi-turn gap stems more from unreliability than reduced aptitude.} The preceding sections report mean performance, but they do not show whether the multi-turn setting consistently lowers performance or instead makes outcomes more variable across repeated runs.
Following the original benchmark, we therefore examine aptitude (90th-percentile F1 across a question's five simulations) and unreliability (difference between 90th and 10th percentiles). We find that under \textsc{HISTORY}, unreliability increases from 13.4 F1 points under \textsc{FULL} to 19.7 under \textsc{SHARDED}, a 47\% increase, while estimated aptitude falls by only 1.9 points. Consistent with the 
original benchmark, 
multi-turn degradation in RAG is thus driven more by unreliability than by reduced aptitude (Appendix~\ref{app:reliability}).

We also test whether agent-style interventions can mitigate this degradation by considering \textsc{RECAP} and \textsc{SNOWBALL}, which consolidate or repeat information from earlier turns (Appendix~\ref{app:mitigation}). Here, the restated information affects not only the assistant but also the retrieval query.
\textbf{We find that for the retrievers least affected by formulation (Section~\ref{sec:concat_decomposition}), consolidation can fully recover
single-turn performance}, while for the remaining retrievers it provides only partial recovery.
Beyond these conversation-side interventions, we also evaluate Plan-on-Graph~\citep{chen2024plan}, an agentic retrieval method in which the LLM plans subquestions and guides graph search. We observe that its multi-turn gap nearly disappears, shrinking to about 2 F1 points (6\%) under \textsc{HISTORY}. However, this ``improvement'' comes with much higher token usage, more LLM calls, and second-lowest overall F1 scores of any retrieval method we test. This suggests that agentic retrieval can reduce multi-turn degradation, but substantial work remains to make it more capable and efficient (Appendix~\ref{app:pog}).

Finally, we address a further concern. Token F1 may overstate the degradation by penalizing lexically different but semantically correct answers. We validate our findings on a three-LLM subset using LLM-as-a-judge evaluation of semantic correctness. Our results show that the degradation remains consistent across retrieval methods and query policies, indicating that the observed gap is not an artifact of lexical-overlap scoring (Appendix~\ref{app:llm-judge}).

\section{Conclusion}
We conduct a large-scale, controlled study of how retrieval-augmented LLMs behave when fully specified benchmark questions are instead spread across a conversation. Across the retrieval methods and LLMs we test, retrieval does not eliminate the \emph{lost in conversation} effect observed in prior work~\citep{laban2026llms}. RAG systems remain less accurate and less reliable across turns. Our decomposition traces this degradation to two sources. Some systems are \emph{lost in translation}, losing performance as soon as the question is reformulated into conversational clues, while others are \emph{lost in conversation}, retrieving the needed evidence but failing to use it across turns. These failures, however, are not inevitable. At least one model retains nearly all of its single-turn performance, and for some retrieval families, consolidating the conversation can fully recover the gap. As retrieval methods 
advance on single-turn benchmarks, our results underscore the need to additionally evaluate them under multi-turn, conversational delivery, so that robustness 
is demonstrated rather than assumed.

\subsection*{Reproducibility statement}
We have taken several steps to ensure reproducibility of our work. We outline
our sharding procedure in Appendix~\ref{app:shard_construct} and disclose all
prompts in Appendix~\ref{app:prompts}. Retriever configurations, model
versions, and experimental methodology are detailed in
Sections~\ref{sec:benchmark}--\ref{sec:setup} and
Appendix~\ref{app:retrievers}. We plan to publicly release our codebase, experiment
results, and project documentation.

\subsection*{AI use statement}
In this work, we use generative AI for four purposes. First, the benchmark is built with a semi-automatic shard construction pipeline that relies on an LLM. Every resulting shard set was manually reviewed and corrected accordingly by the authors before any experiment was run. Second, we use LLM as an automated judge to assess answer correctness for an ablation study (Appendix~\ref{app:llm-judge}). Third, we use AI tools to assist with writing the scripts that generate the paper's figures. Lastly, we also correct grammatical errors in our text with the help of an AI tool. All ideas, experimental design, results and interpretations are the authors' own, and the authors take full responsibility for the paper's content. 
\subsection*{Acknowledgments}
This research was funded in part by DOE grants DE-SC0022098 and DE-SC0023349 and by NSF grants PPoSS CCF 2316233 and OAC-2339607.

\bibliographystyle{iclr2027_conference}
\bibliography{iclr2027_conference}
\newpage
\appendix
\section{Benchmark Construction and Validation}
\label{app:benchmark}
\subsection{Shard Construction and Validation}
\label{app:shard_construct}

We follow the two-stage segmentation and conversationalization procedure of \citet{laban2026llms}. Given a fully specified question, GPT-4o-mini first divides its informational content into non-overlapping segments. A second prompt selects the segment that expresses the user's objective and rewrites it as the opening turn. The remaining segments become successive user turns.

Due to differences in question structure across the three datasets,
we use separate prompt pairs for HotpotQA, 2WikiMultiHopQA, and
MuSiQue (example in Appendix~\ref{app:prompts}). For comparison questions, the prompts separate the requested comparison from the entities being compared and introduce each candidate once. For questions with dependent relations, later turns may refer to entities introduced earlier. In general, we keep each follow-up focused on one unresolved relation at a time. This preserves the underlying reasoning chain without collapsing several dependencies into a single turn or exposing them as explicit subquestions. These and four other constraints are formalized as properties \textbf{P1--P6} below.

These prompts implement a set of six properties that every accepted shard must satisfy. We adopt the information-preservation and surface-fidelity requirements of the original benchmark and refine its initial-intent property. Order insensitivity and maximal sharding, however, are discarded. We find that they do not transfer to multi-hop questions where clues resolve one another, and we replace them with a dependency-ordered revelation framework as mentioned previously in this section. 

\textbf{P1 (Information Equivalence).} The shard set carries exactly the information of the original question.  No relation, entity, date, or value necessary to answer may be dropped, and none may be added -- a shard may not introduce specifics, structure, or causal links that the original question does not state. 

\textbf{P2 (Resolution Blindness).} No shard names the final answer or any intermediate entity that the question only resolves to during reasoning. 

\textbf{P3 (Underspecified Intent).} The first shard states the answer target and the relations through which the answer is found, while withholding every identifying specific. It may not be more informative than the original question's own framing.

\textbf{P4 (Dependency-ordered revelation).} Follow-up shards may bind entities introduced earlier through placeholder references (\emph{``that person is\ldots''}), and shards are revealed in a fixed order consistent with the question's dependency structure.

\textbf{P5 (Single-clue granularity).} Each follow-up introduces one substantive clue: one relation applied to a named anchor or to a single unresolved placeholder. A relation with its named anchor is one clue and is never split. A nested chain is never packed into one. Comparisons introduce each candidate as its own constraint. Turns that add no information are discarded. No specific shard count is targeted. The number of follow-ups emerges naturally from the question's clue structure.

\textbf{P6 (Verbatim preservation).} Shards keep the question's own language. Names, titles, dates, numbers, and unusual phrasing carry over verbatim, and so do the question's defects. Ambiguity or awkward wording present in the original is never clarified or repaired. 

Crucially, these properties control the semi-automatic generation but do not guarantee it. Multi-hop questions ---especially at three- and four-hops---are inherently convoluted, and the sharding LLM can still hallucinate entities, merge separate dependencies into one turn, or pack two clues into a single shard.  We therefore manually reviewed every shard set in the benchmark before any substantial experimental budget was committed. Reviewers edited shards that violated a named property, removed questions whose source text contained factual errors or missing referents, and left the rest unchanged. All sharding prompts are included in our public release.

\begin{figure}[ht]
\centering
\includegraphics[width=\linewidth]{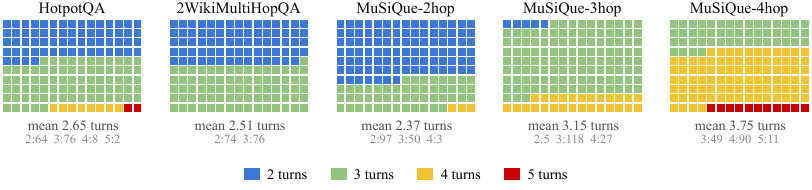}
\caption{Distribution of user turns in the constructed shard sets. Each square is one of the 150 evaluated questions in a dataset.}
\label{fig:shard_dist}
\end{figure}

Table~\ref{tab:shard_examples} shows representative outputs from the three dataset families. Figure~\ref{fig:shard_dist} summarizes the
conversation lengths across the complete benchmark.
\begin{table}[th]
\centering
\caption{Examples of shard construction across the five evaluated datasets. Highlighting marks the span of the \textsc{FULL} question that each user turn carries, and the coloured square on a turn matches its span. The order of the colored spans in the \textsc{FULL} question may differ from the order of the turns below, because shards are revealed according to their dependencies.}
\input{shard_examples.tex}
\label{tab:shard_examples}
\end{table}

One thing worth noting is that turn count reflects how many dependencies must be introduced separately rather than directly reproducing a dataset's annotated hop count.

\subsection{User Simulation and Response Classification}
\label{app:user_sim}

A conversation in our simulation begins with the intent shard
presented verbatim to the assistant as the first user turn. From there, a
fixed GPT-4o-mini user simulator takes over. After each assistant
response, it receives the dialogue so far and the remaining
unrevealed shards, selects at most one, and rewrites it as a short
continuation of the conversation. The prompt requires the simulator to retain all information in the selected shard without drawing on any other unrevealed shard. We use temperature 1.0 and a maximum output length of 200 tokens. Although the next shard is selected in response to the dialogue, we find 98.7\% of conversations retained the dependency order established during construction. This indicates that the simulated conversations largely retain the intended reasoning chain, rather than being contaminated by out-of-order or contextually premature shards.

The information introduced by each shard is fixed, but its wording can vary across simulations. To illustrate this variation, we examine one HotpotQA example for the GPT-4o-mini assistant under BM25, dense RAG, and HippoRAG2 with the \textsc{HISTORY} query policy. Across five simulations of each setting, the opening intent was identical in all 15 conversations as intended. The following census shard appeared in 11 distinct forms, all of which retained the year and the reference to the census. Table~\ref{tab:shard_paraphrases} showcases these variations.

\begin{table}[th]
\centering
\caption{Five realizations of one HotpotQA clue, sampled from
15 conversations with the GPT-4o-mini across 3 different retrieval configurations}
\input{shard_paraphrases.tex}
\label{tab:shard_paraphrases}
\end{table}
As shown in Figure~\ref{fig:simulation_overview}, the simulation also uses a fixed GPT-4o-mini response classifier to categorize the assistant's responses. We adopt the seven response categories of the original benchmark~\citep{laban2026llms}, adapted from \citet{herlihy2024overcoming}:
(1)~answer attempt, (2)~clarification, (3)~interrogation, (4)~discussion,
(5)~hedging, (6)~refusal, and (7)~missing response. Only responses classified as answer attempts are passed to the deterministic answer parser and scored. The classification prompt is provided in Appendix~\ref{app:prompts}.

Answer attempts account for 44.2\% of \textsc{SHARDED} assistant turns with retrieval under \textsc{HISTORY}, followed by clarification requests (40.7\%) and discussion (9.0\%); refusals, interrogations, hedging, and missing responses make up the remaining 6.1\%. The distribution is nearly identical under \textsc{CURRENT} (43.8\%, 40.9\%, and 9.1\%).

\subsection{Closed-Book Comparison}
\label{app:closed_book}
\begin{figure}[th]
\centering
\includegraphics[width=\linewidth]{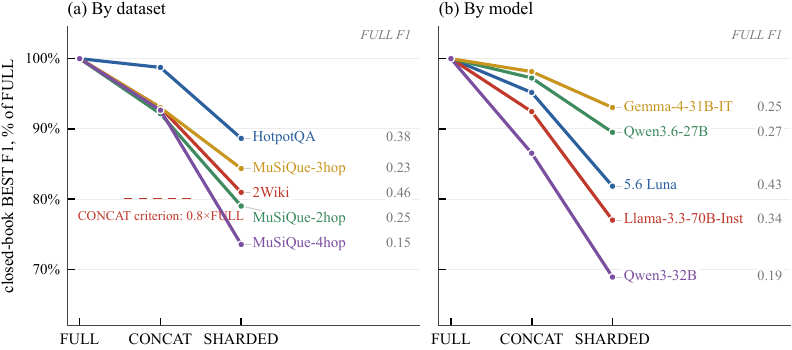}
\caption{Closed-book performance across different simulation conditions.
The displayed models in Panel (b) are 5.6 Luna and the largest evaluated
member of each open-weight model series: Llama-3.3-70B, Gemma-4-31B,
Qwen3.6-27B, and Qwen3-32B.}
\label{fig:closedbook}
\end{figure}

Figure~\ref{fig:closedbook} gives the dataset- and model-level
closed-book results for the five-LLM comparison reported in the main text. Table~\ref{tab:closedbook_appendix} reports the complete results for all ten models.

We observe that the \textsc{CONCAT} degradation is most pronounced
among smaller models. Llama-3.1-8B-Inst and Qwen3-8B both fall below
the 80\% information-preservation criterion, while Gemma-4-31B-IT and
Qwen3.6-27B retain 98\% and 97\%, respectively. This sensitivity to paraphrasing reinforces the observation of the original benchmark that smaller models are particularly vulnerable to
surface-level rephrasing, even when the semantic content of the
question is preserved. Broader studies of LLM sensitivity to paraphrased prompts report the same pattern~\citep{sclar2024quantifying,sun2024evaluating}.

\begin{table}[th]
\centering
\small
\setlength{\tabcolsep}{4pt}
\caption{Complete closed-book results. F1 is
reported on a 0--1 scale.}
\input{closedbook_appendix.tex}
\label{tab:closedbook_appendix}
\end{table}

\newpage
\section{Additional Results and Analyses}
\label{app:additional}
\subsection{Agent-Style Consolidation: RECAP and SNOWBALL}\label{app:mitigation}

If the multi-turn gap were caused only by information arriving in pieces, then reassembling the pieces should remove it. \citet{laban2026llms} test this with two agent-style interventions and find that both improve multi-turn performance without restoring the single-turn baseline. We repeat both interventions with retrieval in the loop. \textsc{RECAP} appends one final turn to the conversation that consolidates all previously revealed shards and asks the model to answer again. \textsc{SNOWBALL}, in contrast, repeats all previously revealed shards at every turn. We run both conditions on three LLMs (GPT-4o-mini, Qwen3-32B, and Llama-3.3-70B) across all five datasets and seven retrieval settings. The two interventions differ in how they handle the end of the conversation. \textsc{RECAP} assumes that the final turn is known in advance, which is realistic only in a simulation-like setting. \textsc{SNOWBALL}, by contrast, can be applied throughout the interaction without knowing when the conversation will end, making it the more practical intervention.

It is worth noting that the retrieval query policy also changes what
these interventions present to the retriever. Under \textsc{HISTORY}, the recap-turn query contains every shard twice: once in its original conversational turn and once in the final restatement. For \textsc{SNOWBALL}, earlier shards are repeated increasingly often as the conversation grows. Under \textsc{CURRENT}, the recap query consists only of the final restatement, which contains the same shard content as \textsc{CONCAT} together with the recap framing sentence. We therefore treat \textsc{CURRENT} as the standard test of consolidation, since each shard appears exactly once in the retrieval query, while using \textsc{HISTORY} as a complementary setting.
\begin{figure}[th]
\centering
\includegraphics[width=\linewidth]{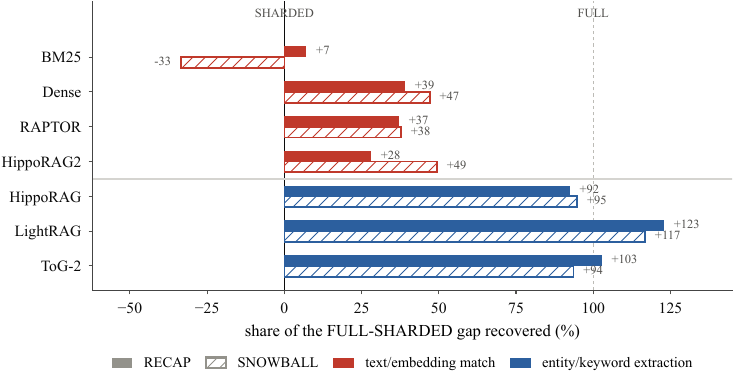}
\caption{Recovery of the \textsc{FULL}--\textsc{SHARDED} performance gap
under \textsc{RECAP} and \textsc{SNOWBALL}, using the \textsc{CURRENT}
retrieval query. Recovery is
normalized such that 0\% corresponds to \textsc{SHARDED} and 100\% to
\textsc{FULL}. Negative values indicate performance below
\textsc{SHARDED}.}
\label{fig:mitigation_recovery}
\end{figure}

\begin{figure}[h]
\centering
\includegraphics[width=\linewidth]{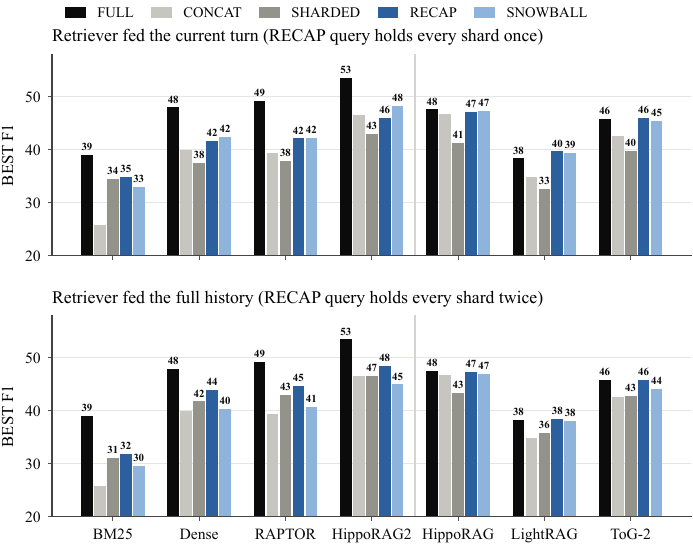}
\caption{Mean F1 ($\times 100$) under all five conditions, averaged over
the three LLMs and five datasets. The upper panel uses
\textsc{CURRENT} retrieval queries and the lower panel uses
\textsc{HISTORY}. Colors distinguish the two consolidation interventions.}
\label{fig:mitigation}
\end{figure}

We first consider \textsc{RECAP} under \textsc{CURRENT}. Figure~\ref{fig:mitigation_recovery} shows \textbf{consolidation is highly effective for HippoRAG, LightRAG, and ToG-2}, recovering 92--123\% of the \textsc{FULL}--\textsc{SHARDED} gap and bringing performance back to approximately the \textsc{FULL} level. BM25, Dense, RAPTOR, and HippoRAG2 recover much less, at 7--39\%. This division closely follows the formulation losses in Table~\ref{tab:gap_decomposition}: HippoRAG, LightRAG, and ToG-2 lose only 1.4--2.8 F1 points from \textsc{FULL} to \textsc{CONCAT}, whereas the remaining retrievers lose 7.1--11.7 points. The pattern therefore echoes the earlier decomposition: consolidation works best for retrieval methods that are already relatively insensitive to the sharded formulation. The same qualitative division holds under \textsc{HISTORY} (Figure~\ref{fig:mitigation}).

To understand why, we examine the evidence retrieved at the recap turn itself. We observe that the recall shows the same separation as the answer scores. HippoRAG retrieves at least as much supporting evidence under \textsc{RECAP} as it does under \textsc{FULL}, whereas BM25 remains below both \textsc{CONCAT} and its own final \textsc{SHARDED} retrieval. The consolidated restatement contains all of the question information, yet BM25 still retrieves less useful evidence from it than from the final \textsc{SHARDED} turn. This behavior mirrors the formulation split from
Section~\ref{sec:concat_decomposition}: BM25, Dense, RAPTOR, and
HippoRAG2 operate directly on the supplied query text or its
embedding, whereas HippoRAG, LightRAG, and ToG-2 first transform
the query into entities or keywords. The latter group is therefore less dependent on the exact surface form of the consolidated question, which helps explain why \textsc{RECAP} restores much more of their answer performance.

\textsc{SNOWBALL} shows a similar pattern under \textsc{CURRENT}. Because each new turn carries forward all previously revealed shards, the retriever receives a progressively more complete query as the interaction unfolds. HippoRAG, LightRAG, and ToG-2 recover 94--117\% of the gap, while Dense, RAPTOR, and HippoRAG2 recover 38--49\%. BM25 is again the exception and remains below plain \textsc{SHARDED}. Repeating earlier information can therefore recover substantial performance, but only when the resulting turn also forms an effective retrieval query.

Overall, the two interventions point to the same conclusion. Restating earlier turns can recover most, if not all, of the multi-turn gap, going beyond the partial recovery reported by the original benchmark in a non-retrieval setting. The extent of this recovery, however, depends strongly on the retriever. Methods that first extract entities from the query to start their graph-retrieval procedure return close to their single-turn performance, whereas methods that rely directly on the surface form of the text recover only part of that gap. Consolidation therefore helps a RAG system only when the resulting restatement also serves as an effective retrieval query.

\subsection{Adaptive Graph Planning Retrieval: Plan-on-graph}
\label{app:pog}

The consolidation conditions in the previous section restate the question \emph{before} retrieval. A different approach is to let the LLM plan the retrieval process extensively. Plan-on-graph (PoG) \citep{chen2024plan} is a self-correcting adaptive planning method for knowledge-graph augmented LLMs. Given a query, PoG first uses the LLM to decompose it into sub-objectives. It then explores the graph one hop at a time. At each hop, the LLM has to select which relations to follow, which resulting entities to prune. It also updates a running memory of the explored subgraph, the reasoning paths found so far, and the status of each sub-objective. And after each hop, the LLM judges whether the accumulated evidence is enough for a correct answer. If it is not, a reflection step can add new starting entities or backtrack to an earlier one. This process of exploration continues until the evidence is considered sufficient or the maximum depth of four hops is reached. 

ToG-2~\citep{ma2025think} also uses the LLM to guide graph traversal but with far tighter constraints and less LLM usage. It makes one batched LLM call per hop to select relations for a beam of at most three entities (width $W{=}3$), ranks the resulting entities with dense passage scores and stops after a fixed maximum depth matched to each dataset's hop count. Unlike PoG, ToG-2 has no memory or reflection step. Our PoG adaptation runs over the same corpus-derived OpenIE graph as ToG-2 and the other graph retrievers. Because of the high per-conversation cost of PoG's planning loop, we evaluate it on the eight open-weight models we use in the main experiments. All results reported in this section are averaged over these eight LLMs and the five datasets unless stated otherwise.

\begin{figure}[th]
\centering
\includegraphics[width=\linewidth]{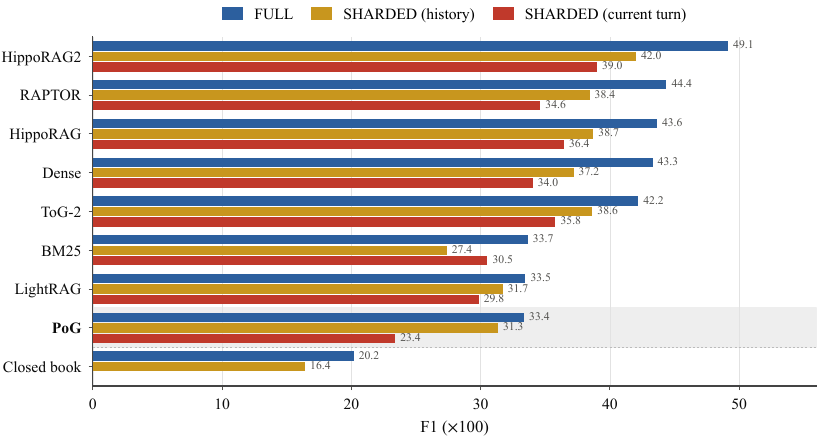}
\caption{Averaged F1 ($\times 100$) under \textsc{FULL} and both \textsc{SHARDED} policies. Rows are ordered by \textsc{FULL} score.}
\label{fig:pog_f1}
\end{figure}

As PoG decomposes the query internally, it follows that its retrieval should benefit from seeing the accumulated question history, i.e., the \textsc{HISTORY} policy. We observe exactly this in our results (Figure~\ref{fig:pog_f1}). Under \textsc{HISTORY}, PoG loses only 2.0 F1 points (6.1\%) between \textsc{FULL} and \textsc{SHARDED}; on HotpotQA, \textsc{SHARDED} even performs better than \textsc{FULL} (50.4 vs.\ 49.7). Under \textsc{CURRENT}, however, the loss grows to 10.0 points (29.9\%), the largest relative decline of any retrieval method. This indicates that, with only the latest shard as its query, PoG has too little of the question to decompose effectively.

One thing worth noting is that, PoG's small \textsc{HISTORY} gap starts from a low absolute level. PoG's \textsc{FULL} score of 33.4 is level with BM25 (33.7) and LightRAG (33.5), the two weakest retrieval methods. Under \textsc{SHARDED} with \textsc{HISTORY}, PoG ranks second to last among all the retrieval methods. 

These results do not support adaptive graph planning as a practical alternative to simpler methods for reducing multi-turn degradation. PoG's small \textsc{HISTORY} gap comes with low single-turn accuracy, the largest \textsc{CURRENT} loss of any method, and substantially higher retrieval cost. Each \textsc{FULL} retrieval requires 22.1 LLM planning calls, compared with 2.5 for ToG-2. This difference is not explained by depth alone, since on MuSiQue 4-hop, both methods may explore four hops and PoG still makes $9.0\times$ as many calls. Over a full \textsc{SHARDED} conversation, PoG makes 47.5 LLM calls for retrieval alone, against 6.9 for ToG-2. Each PoG call is short, so the token overhead is $2.9\times$ that of ToG-2. Unlike the conversation-side consolidation in the previous section, where
restating the question recovers most of the gap for some retrievers, PoG's retrieval-side planning does not translate its additional LLM usage into competitive answer quality.

\subsection{Aptitude and Unreliability}\label{app:reliability}
The results so far report mean scores, but as noted in
Section~\ref{sec:metrics}, repeated simulations of the same
configuration can produce different answers. A mean score can
therefore mask whether multi-turn delivery makes a system uniformly
worse or merely less predictable. To quantify this, we adopt the
aptitude and unreliability measures of the original
benchmark~\citep{laban2026llms}. For each question, let
$\mathbf{S} = (S_1,\ldots,S_5)$ denote the F1 scores from five
simulations with the assistant, retrieval method, query policy, and
condition fixed. We define aptitude $A$ and unreliability $U$ as
\begin{equation}
\begin{aligned}
A &= \operatorname{percentile}_{90}(S), \\
U &= \operatorname{percentile}_{90}(S)
   - \operatorname{percentile}_{10}(S).
\end{aligned}
\label{eq:aptitude_unreliability}
\end{equation}

We use linear interpolation to estimate the percentiles.
These are descriptive estimates from five simulations.
We then average the per-question values across datasets and
the models or retrieval methods being compared.

\begin{figure}[thbp]
\centering
\includegraphics[width=\linewidth]{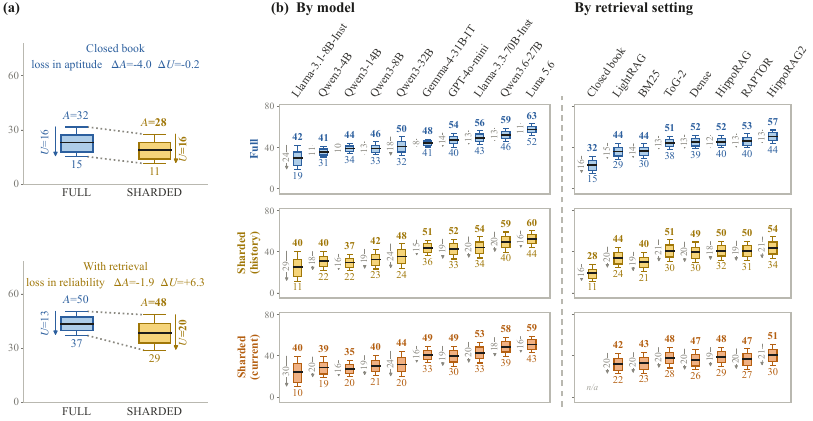}
\caption{Aptitude and unreliability under multi-turn setting.
(a)~Without retrieval, aptitude falls while unreliability remains
the same. With retrieval, aptitude holds but unreliability increases
sharply. (b)~Per-model and per-retriever breakdown, ordered by mean
\textsc{FULL} score.}
\label{fig:aptitude_reliability}
\end{figure}

Figure~\ref{fig:aptitude_reliability} shows the comparison. Without retrieval, both the ceiling and the floor fall by similar
amounts (4.0 and 3.8 points), so the distribution shifts down as a
whole and unreliability is unchanged. With retrieval, however, the ceiling falls only 1.9 points while the floor drops 8.2, and unreliability widens from 13.4 to 19.7. In other words, closed-book multi-turn loss is a
uniform decline in performance, whereas retrieval-augmented loss
concentrates in the lower-scoring simulations and leaves the best
outcomes largely intact.
\begin{table}[h]
\centering
\caption{Aptitude $A$, unreliability $U$, and mean F1 $P$
($\times 100$) under \textsc{FULL} and \textsc{SHARDED}
(\textsc{HISTORY}). Shading marks the change in $U$.}
\input{reliability_levels.tex}
\label{tab:reliability_levels}
\end{table}

Under \textsc{HISTORY}, unreliability increases for all ten LLMs under \textsc{SHARDED}, with gains ranging from 5.1 to 7.8 points.
Aptitude falls for eight of the ten. The two exceptions are
Gemma-4-31B, whose aptitude actually improves by 3.1 points, and
Qwen3.6-27B, which  essentially stays the same (+0.3). Gemma is a
particularly revealing case. On one hand, it remains the most
consistent model even after sharding, with the lowest \textsc{SHARDED} unreliability of all ten LLMs (15.0). On the other
hand, its unreliability nearly doubles from its \textsc{FULL} value
of 7.6, even though its mean F1 drops only 0.7 points
(Section~\ref{sec:not_eliminate}). The small mean change hides the
fact that its best runs improved while its worst runs deteriorated.
\textbf{A model's mean robustness to multi-turn conversation can conceal
an increase in unreliability.}

We observe the same pattern under \textsc{CURRENT}, where mean
unreliability reaches 19.9 points. Every model and every retrieval
method shows higher unreliability than under \textsc{FULL} with
either query policy. Table~\ref{tab:reliability_levels} reports the
\textsc{HISTORY} values.

\subsection{Answering under Complete Supporting-Source Coverage}
\label{app:evidence_coverage}

Section~\ref{sec:retrieval_coverage} showed that cumulative retrieval coverage under \textsc{SHARDED} approaches or exceeds the \textsc{FULL} baseline while answer quality remains lower. Both observations are averages, however, and could in principle describe different conversations: the ones that retrieve the evidence need not be the ones which answer poorly. In order to assess whether the finding holds at the level of individual conversations, we conduct a paired comparison of \textsc{FULL} and \textsc{SHARDED} under \textsc{HISTORY}. Each pair shares the LLM, dataset, question, retriever and simulation. We consider only those pairs whose conversations retrieve all of the question's supporting passages. One thing worth noting is that complete supporting-source coverage does not remove distractor passages from the retrieved context. To answer correctly, the LLM must therefore identify the relevant evidence among these distractors and combine it across turns to produce the final answer.

Among the 54,876 qualifying pairs (24.4\% of the 225,000 eligible), we find that mean BEST F1 falls from 73.7 under \textsc{FULL} to 62.1 under \textsc{SHARDED}, \textbf{a difference of 11.6 points} (95\%
paired bootstrap CI [9.9--13.4]). This effect is not confined to any single dataset: Table~\ref{tab:evidence_coverage_datasets} shows significant declines on four of the five datasets. Moreover, complete coverage becomes increasingly rare as the required evidence grows, falling from 46.9\% of HotpotQA pairs to 0.5\% on MuSiQue-4hop. This sharp decline highlights a practical limitation of current retrieval methods: as more supporting passages are required, retrieving all of them becomes increasingly difficult. 

\begin{table}[th]
\centering
\small
\setlength{\tabcolsep}{6pt}
\caption{Paired F1 under complete supporting-source coverage across datasets (\textsc{HISTORY}, six retrievers, RAPTOR excluded).
$\bullet$: the paired bootstrap 95\% interval of the gap lies above zero.}
\input{evidence_coverage_datasets.tex}
\label{tab:evidence_coverage_datasets}
\end{table}

\begin{figure}[t]
\centering
\includegraphics[width=0.62\linewidth]{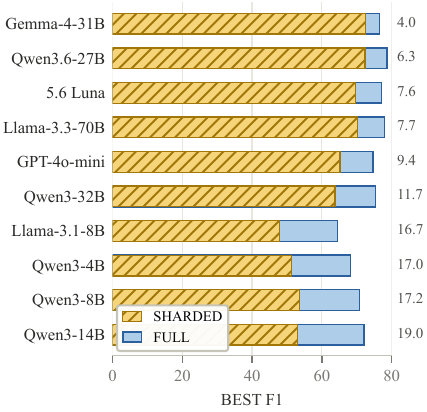}
\caption{Paired F1 by LLM under complete supporting-source
coverage. \textsc{FULL}--\textsc{SHARDED} difference reported on the right.}
\label{fig:evidence_coverage_llm}
\end{figure}

Looking across LLM assistants, however, we observe that the 11.6-point aggregate gap is far from uniform across models. Figure~\ref{fig:evidence_coverage_llm} reveals pronounced model-level differences: every LLM retains a significant decline, but the magnitude varies almost fivefold, from 4.0 F1 points for Gemma-4-31B to 19.0 for Qwen3-14B. Six models lose 4-12 points, while the four smallest lose 17--19 points and retain only about three quarters of their \textsc{FULL} performance. \textbf{Encouragingly, when all supporting sources are retrieved, the stronger LLM assistants preserve substantially more of their single-turn performance than the smaller models.} The most robust retain over 90\% of their \textsc{FULL} performance, suggesting that large multi-turn losses are not inevitable, although the remaining gap leaves clear room for improvement.

These conclusions, however, apply only to the 24.4\% of paired conversations in which both conditions retrieve every supporting passage. In the remaining pairs, \textsc{FULL} and \textsc{SHARDED} can also differ in how much supporting evidence they retrieve, so retrieval itself may contribute to the overall loss, as examined in Sections~\ref{sec:concat_decomposition} and~\ref{sec:retrieval_coverage}.

\subsection{LLM-as-a-Judge Evaluation}
\label{app:llm-judge}

Token F1 rewards lexical overlap and penalizes semantically equivalent answers that differ in wording from the reference answer \citep{kamalloo2023evaluating}. We therefore test whether the \textsc{FULL} - \textsc{SHARDED} degradation persists under binary semantic-correctness measures. We follow prior QA and RAG work that use an LLM to assess a candidate answer against the question and reference answer \citep{rau2024bergen}.

We rescore the conversations for Llama-3.3-70B, Gemma-4-31B, and
Qwen3-32B across all five datasets, all retrieval methods with both
query policies, and the closed-book setting. We use GPT-4o-mini at
temperature zero to assign binary
\textbf{CORRECT}/\textbf{INCORRECT} judgments. For each stored answer
attempt, the LLM receives the original question, the reference answer
and its aliases, and the extracted candidate answer. The judge is
instructed to accept semantically equivalent answers despite lexical
or formatting differences, while rejecting contradictory or ambiguous
responses. We apply the BEST convention from
Section~\ref{sec:metrics} and compute paired 95\% question-cluster bootstrap confidence intervals using 5,000 resamples, keeping all observations associated with each question together.

\begin{figure}[t]
\centering
\includegraphics[width=0.9\linewidth]{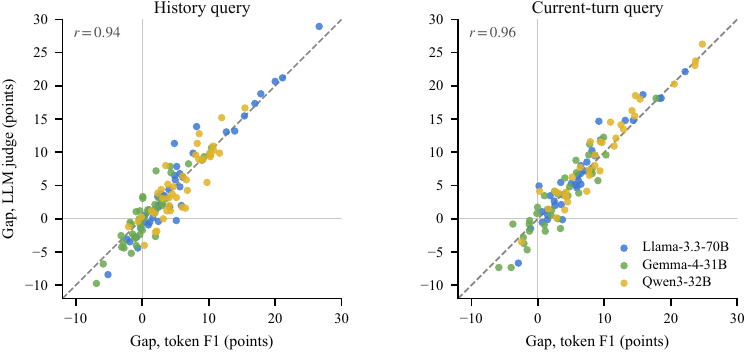}
\caption{\textbf{\textsc{FULL}--\textsc{SHARDED} gaps are consistent across metrics.} Each point represents one model--dataset--retriever combination, comparing the gap under token F1 with the same gap under the binary judge.}
\label{fig:judge_robustness}
\end{figure}

Table~\ref{tab:judge_levels} and
Figure~\ref{fig:judge_robustness} compare the two metrics. We find
that the mean \textsc{HISTORY} gap changes only from 4.1 points under
token F1 to 3.8 under the judge, and every retrieval method retains a
positive mean gap under both query policies. The per-configuration gaps under token F1 and the LLM judge are
strongly correlated, with $r = 0.94$ under \textsc{HISTORY} and
$0.96$ under \textsc{CURRENT}. Directly comparing the two measured gaps, the token-F1 gap exceeds the judge gap by 0.3 points under \textsc{HISTORY} (95\% CI $[-0.2, 0.8]$) and is 0.2 points smaller under \textsc{CURRENT} (95\% CI $[-0.8, 0.5]$). Both intervals include zero, and no gap reverses from a significant decline to a significant improvement under the bootstrap intervals.

\begin{table}[t]
\centering
\caption{F1 scores
($\times$100) and binary judge accuracy under the \textsc{HISTORY} query policy, pooled over Llama-3.3-70B,
Gemma-4-31B, and Qwen3-32B and the five datasets. Shading marks the size of the gap.}
\input{judge_levels.tex}
\label{tab:judge_levels}
\end{table}

Because LLM judges can carry their own
biases~\citep{zheng2023judging}, we treat this analysis as a
robustness check rather than a replacement for F1. That said,
the \textsc{FULL}--\textsc{SHARDED} degradation remains nearly
unchanged under a judge explicitly instructed to accept semantically
equivalent wording, confirming that the observed loss reflects answer
quality rather than lexical mismatch.

\section{Qualitative Retrieval and Conversation Example}
Figure~\ref{fig:conversation_example} shows a representative \textsc{SHARDED} conversation, in which the assistant commits to an incorrect date before arriving at the correct answer at a later turn.
\begin{figure}[th]
\centering
\input{conversation_example.tex}
\caption{A \textsc{SHARDED} conversation with Llama-3.3-70B and HippoRAG
under \textsc{CURRENT} retrieval queries.}
\label{fig:conversation_example}
\end{figure}

\section{Retrieval Systems and Configuration Summary}
\label{app:retrievers}
We evaluate retrieval families spanning lexical passage retrieval, dense passage retrieval, knowledge-graph traversal, and hierarchical retrieval. The descriptions below summarize the evaluated systems; implementation details are reported in Table~\ref{tab:retriever_implementation}.

\subsection{Retrieval Systems}
\begin{itemize}
\item \textbf{NONE.} A closed-book control that measures the assistant's parametric performance without external evidence.
\item \textbf{BM25}~\citep{robertson2009probabilistic}. A lexical passage retriever that favors exact term
overlap and returns the five highest-ranked passages from the shared corpus.
\item \textbf{Dense TextRAG}~\citep{xiao2024c}. A semantic passage retriever that ranks the
shared corpus with normalized BGE-large embeddings and returns five passages.
\item \textbf{HippoRAG}~\citep{gutierrez2024hipporag}. A memory-inspired graph retriever that links query entities to a knowledge graph (NER) and uses Personalized PageRank to propagate relevance across related concepts before ranking supporting passages.
\item \textbf{HippoRAG2}~\citep{gutierrez2025rag}. A graph retriever that combines
phrase-level conceptual structure with passage-level context and uses language-model recognition to
filter candidate facts.
\item \textbf{RAPTOR}~\citep{sarthi2024raptor}. A hierarchical retriever that recursively organizes
passages into a tree of leaves and summaries, allowing retrieval at different levels of abstraction.
\item \textbf{ToG-2}~\citep{ma2025think}. A hybrid graph-text retriever that tightly couples knowledge-graph traversal with passage retrieval. It uses an LLM to select promising graph relations and reason over retrieved evidence while passage relevance guides subsequent graph exploration.

\item \textbf{LightRAG}~\citep{guo2025lightrag}. A graph-enhanced retriever that extracts local and global query keywords. Uses vector search to match them to entities and relationships, and uses graph connectivity to retrieve additional relevant evidence.
\end{itemize}

\subsection{Implementation Details}
Table~\ref{tab:retriever_implementation} summarizes the indexing, retrieval, and generation
interfaces used in the benchmark.

\begin{table}[h]
\caption{Implementation details of the evaluated retrieval families.}
\label{tab:retriever_implementation}
\scriptsize
\resizebox{\linewidth}{!}{%
\begin{tabular}{@{}l|ll|ll|l@{}}
\toprule
\textbf{Retriever} & \multicolumn{2}{c|}{\textit{Indexing}} & \multicolumn{2}{c|}{\textit{Retrieval}} & \textit{Generation} \\
 & \textbf{Knowledge type} & \textbf{Index content} & \textbf{Query input} & \textbf{Granularity} & \textbf{Model context} \\
\midrule
NONE & Closed-book & --- & --- & --- & Parametric only \\
BM25~\citep{robertson2009probabilistic} & Plain text & Frozen passages & Lexical terms & Passage & 5 retrieved passages \\
Dense TextRAG~\citep{xiao2024c} & Plain text & Passage embeddings & Dense query & Passage & 5 retrieved passages \\
HippoRAG~\citep{gutierrez2024hipporag} & Phrase graph & Phrases, passages & Query phrases & Entity, passage & 5 retrieved passages \\
HippoRAG2~\citep{gutierrez2025rag} & Fact graph, passages & Facts, entities, passages & Query-to-fact & Fact, entity, passage & 5 retrieved passages \\
RAPTOR~\citep{sarthi2024raptor} & Summary tree & Passage leaves, summaries & Dense query & Collapsed-tree node & 5 leaf/summary nodes \\
ToG-2~\citep{ma2025think} & OpenIE graph, passages & Entities, relations, passages & Query entities & Path, passage & Triples + 5 passages \\
LightRAG~\citep{guo2025lightrag} & Entity--relation graph, text & Entities, relations, chunks & Native keywords & Local, global & Entities, relations, passages \\
\bottomrule
\end{tabular}}
\end{table}

\subsection{Parent Corpora}
\label{app:corpora}
Table~\ref{tab:corpus_stats} reports the size of the parent corpus built for each dataset.

\begin{table}[ht]
\centering
\caption{Parent corpora used for retrieval.}
\input{corpus_stats.tex}
\label{tab:corpus_stats}
\end{table}
\newpage
\subsection{Model Versions and Access}
\label{app:model_access}

Table~\ref{tab:model_versions} lists the models used in our
experiments together with their versions and access details. We serve
open-weight models locally with vLLM on NVIDIA A100 GPUs. All models
use temperature 1.0 and a 1,000-token answer cap. We disable reasoning and thinking for all models except two, both of which use a 10,000-token cap. One exception is 5.6~Luna, which reasons natively each turn because its responses frequently exceeded the standard limit. The other is Qwen3.6-27B, which keeps thinking disabled but uses the official non-thinking instruct sampler at temperature 0.7.
The user simulator uses $T$\,=\,1.0 and the response classifier
uses $T$\,=\,0 throughout all experiments.

\begin{table}[h]
\centering
\caption{Evaluated LLMs with versions and access details.}
\label{tab:model_versions}
\input{model_versions.tex}
\end{table}

\subsection{Retriever Configurations}
The principal configurations are given below in a compact form. All systems use the shared corpus and the same answer-generation interface. Retriever-specific settings define only how that corpus is indexed and how evidence is selected.

\begin{configbox}{BM25 Configuration}
retrieval_top_k: 5
\end{configbox}

\begin{configbox}{Dense TextRAG Configuration}
embedding_model: BAAI/bge-large-en-v1.5
similarity: cosine
retrieval_top_k: 5
\end{configbox}

\begin{configbox}{HippoRAG Configuration}
query_linking: phrase_to_entity
graph_propagation: personalized_pagerank
retrieval_top_k: 5
\end{configbox}

\begin{configbox}{HippoRAG2 Configuration}
query_linking: dense_concept_linking
fact_candidates: 5
passage_node_weight: 0.05
retrieval_top_k: 5
\end{configbox}

\begin{configbox}{RAPTOR Configuration}
index_structure: collapsed_tree
retrieval_top_k: 5
summary_length_limit: 100
context_limit: 3500
\end{configbox}

\begin{configbox}{ToG-2 Configuration}
relation_selection: batched
beam_width: 3
seed_entities: 5
retrieval_top_k: 5
traversal_depth: dataset_matched
\end{configbox}

\begin{configbox}{LightRAG Configuration}
query_type: local_and_global
chunk_token_size: 1200
chunk_overlap_token_size: 100
local_evidence_tokens: 400
global_evidence_tokens: 400
\end{configbox}

\newpage
\section{Prompts}
\label{app:prompts}
We disclose the prompts used across our pipeline below. We show the shard-construction prompts for a single dataset (HotpotQA) here. The prompts for the remaining datasets follow the same structure and will be provided in our code repository.

\subsection{Shard Construction}
Each fully specified question is first segmented into information units, and the
segments are then rephrased into a multi-turn conversation. Both steps are run with
the shared GPT-4o-mini scaffold.

\begin{promptbox}[Segmentation]{shardgreen}
{\footnotesize
\begin{Verbatim}[breaklines,breaksymbol=]
You are given a fully specified factual question, and your task is to segment the question into units of information that each represent a single piece of information needed to answer it.

You must output a list of segments in the following JSON format:
{"segments": [
    {"segment": "[short excerpt from the question]"},
    {"segment": "[short excerpt from the question]"}
]}

Rules:
- [Non-overlapping] Segments must be non-overlapping and together cover the entire question. You can leave gaps for non-essential portions (delimiters, articles like "the", question marks).
- [Coherent units] Each segment should be a coherent clue or constraint. Do not split noun phrases, relations, dates, titles, or named entities into fragments. Prefer 2-4 segments for short questions under 30 words. Only use 5+ segments when the question truly contains 5+ independent constraints.
- [Preserve bridge relations] HotpotQA questions often ask through a bridge entity. Keep relation phrases intact, such as "creator of what 2012 film", "album released in 1981", or "grandfather to the children of Mark Antony". Do not reduce them to vague fragments like "what film", "the company", "details", or "connection".
- [No external information] Only segment what's actually in the question. Do not add facts from your own knowledge.

Example Question:
Glenn Quagmire is voiced by the creator of what 2012 film?

Example Output:
{"segments": [
    {"segment": "Glenn Quagmire is voiced by"},
    {"segment": "the creator of what 2012 film"}
]}

Now segment the following question. Output ONLY a JSON object with the "segments" field.

Question: {question}
\end{Verbatim}
}
\end{promptbox}

\begin{promptbox}[Rephrasing]{shardgreen}
{\footnotesize
\begin{Verbatim}[breaklines,breaksymbol=]
You are given segments of a fully specified question. Your task is to turn them into a multi-turn conversation: (1) choose an "intent shard" that opens the conversation by naming WHAT THE USER IS LOOKING FOR, and (2) rephrase each remaining segment into a conversational follow-up turn that adds one clue.

Output a JSON object in the following format:
{
    "answer_target": "[what the question is ultimately asking for: a prize, a film, a date, a person's middle name, a city, etc.]",
    "initial_segment": "[exact segment text the intent shard is based on]",
    "initial_shard": "[short conversational opener naming the answer target - like a user starting a search]",
    "shards": [
        {"segment": "[segment text]", "shard": "[conversational follow-up turn that adds this clue]"}
    ]
}

Rules:
- [Intent shard names the GOAL, not the subject] The initial_shard must announce WHAT THE USER WANTS TO FIND (the answer target), not just the first noun in the question. If the question asks "...won what prize?", the opener is about finding a prize ("trying to find what award someone won"), NOT "an American playwright" (that's a clue, it goes in a follow-up). If the question asks "what 2012 film?", the opener is about a film. If it asks "the middle name of the actress", the opener is about finding a middle name. The conversation should open the way a real user opens: stating their goal.
- [Transform every segment] Every segment must appear exactly once - as the initial_segment or in the shards list. Do not drop or merge segments.
- [Conversational turns] Each shard should sound like a real user speaking a turn - natural, brief, can include filler ("oh and", "btw", "I think"). 1-2 sentences. This is a conversation, not a list of fragments.
- [No new information / no answer leakage] Only re-present what is in the segments. Do NOT add facts from your own knowledge, and never reveal or hint at the actual answer.
- [PRESERVE THE CHAIN - most important] These questions often reach the answer through a chain of hops: an entity, a relation to a bridge entity, then the final thing being asked. Every link MUST survive across the shards, and the final thing being asked must stay the final thing being asked. Do NOT collapse the chain to the bridge entity. Example: "Glenn Quagmire is voiced by the creator of what 2012 film?" must NOT become "what film is Quagmire from" - that drops the "voiced by the creator of" hops and asks a different question.
- [Back-reference is allowed] To keep the chain, a follow-up shard MAY refer back to something in an earlier shard ("that person", "that film", "the same director"). A real multi-turn user speaks this way. Preserving the relation matters more than making each shard independently self-contained.
- [Shards are CONSTRAINTS, not sub-questions] This keeps the shards faithful to a lazy, underspecified user. Each follow-up states a CONSTRAINT or CLUE the user knows about what they want - it does NOT pose a reasoning step for the solver. The user supplies clues; the user does NOT hand over the solution plan. (The intent shard naming the goal is fine - that is the user stating what they want, not a reasoning step.)
      GOOD (clue the user asserts):  "and it was created by whoever voices Glenn Quagmire", "the one directed by that same person", "it came out in 2012"
      BAD  (sub-question that decomposes the task):  "who voices Glenn Quagmire?", "what did that person create?", "first find the director"
  The GOOD form leaves the reasoning to the solver; the BAD form does the solver's job. Phrase follow-ups as declarative clues, not as step-by-step questions for the assistant.
- [Order of shards] Order the follow-up shards from most to least important.

Worked example A - goal-first opener, target named (NOT hinted):
Question: "Venus is a play by an American playwright who won what prize?"
Output:
{
    "answer_target": "a prize / award won by the playwright",
    "initial_segment": "won what prize",
    "initial_shard": "I'm trying to figure out what prize someone won",
    "shards": [
        {"segment": "an American playwright", "shard": "they're an American playwright"},
        {"segment": "Venus is a play by", "shard": "and they wrote the play Venus"}
    ]
}

Worked example B - bridge chain preserved, follow-up is a declarative clue (not a sub-question):
Question: "Glenn Quagmire is voiced by the creator of what 2012 film?"
Output:
{
    "answer_target": "a 2012 film",
    "initial_segment": "the creator of what 2012 film",
    "initial_shard": "looking for a 2012 film",
    "shards": [
        {"segment": "Glenn Quagmire is voiced by", "shard": "it was created by whoever does the voice of Glenn Quagmire"}
    ]
}

Now complete the task for the following question and segments.

Question: {question}

Segments:
{segments_json}
\end{Verbatim}
}
\end{promptbox}

\subsection{Simulation Scaffold}
During each simulated conversation, a shared GPT-4o-mini plays the user (revealing
one shard per turn) and classifies each assistant turn into one of seven response
categories.

\begin{promptbox}[User simulator]{simblue}
{\footnotesize
\begin{Verbatim}[breaklines,breaksymbol=]
You are simulating a user of an interactive LLM system (like ChatGPT).
The user is inherently lazy, and answers in short form, providing only minimal information to the system. You should not be proactive.

Here's the conversation so far:
{conversation_block}

Here are the shards that have already been revealed:
{revealed_block}

Here are all the shards that have not been revealed yet:
{remaining_block}

You must generate a response to the conversation so far. Here are the rules:
- [Providing a Shard] You can reveal the content of a shard to the system in your response if it will help the system move closer to answering the problem. You should select the shard to reveal that is most "basic" and is the current most relevant shard.
- [One Shard at a Time] You should only reveal at most one shard at a time.
- [Never Mix Unrevealed Shards] After selecting one shard, that selected shard is the ONLY source of new information for your response. You may use the conversation and already revealed shards to make the response coherent, but you must not add, combine, hint at, or paraphrase information from any other unrevealed shard. This remains true even when another unrevealed shard is closely related or would make the response more useful.
  Example: suppose unrevealed shard 2 says "and they wrote the play Venus" and unrevealed shard 3 says "they're an American playwright". If you select shard 2:
  WRONG: "They're an American playwright and wrote the play Venus." This leaks information from shard 3.
  CORRECT: "They wrote the play Venus." This conversationally rephrases shard 2 and reveals nothing from shard 3.
- [Preserve the Original Goal] The initial intent shard establishes what the user ultimately wants. A later non-intent shard adds a constraint that identifies or qualifies that original target; it does not replace the original goal with a new intermediate question. Do not introduce a later shard with phrases such as "I'm looking for", "I want to know", or "I'm trying to find out" unless that selected shard is itself the intent shard. State the later constraint declaratively and connect it to the target already under discussion.
  Example: the initial intent is "I'm trying to find out what year a certain Governor ended their term" and the selected later shard is "the city is where the author of Pacem in Terris died".
  WRONG: "I'm looking for the city where the author of Pacem in Terris died." This changes the goal into answering the intermediate city question.
  CORRECT: "The city I mean is where the author of Pacem in Terris died." This conversationally adds the constraint while preserving the original goal.
- [Reveal Entire Shard] If you reveal a shard, you must make sure to include *all the information in the shard*. For example, if the shard is "your symptoms are that you have a headache in the mornings", your response can't just be "yeah I have headaches", you must say "yup mostly headaches in the mornings".
- [Irrelevant Clarifications] If the system asks you a question irrelevant to the shards, asks you a generic question ("Can you give me a hint?"), you should respond with an answer that does not provide a shard. ("I don't know", "Is that really important?", etc.) You should not reveal any information beyond what is available in the shards.
- [No Repeated Shards] You should not reveal the same shard more than once. Carefully review the shards revealed already, and only reveal a shard if its `shard_id` is not on the list.
- [Rephrase Shards] If you reveal a shard, you should rephrase it in a conversational way. Do not copy the shard verbatim.
- [Do Not Ask Questions] Your response should always be declarative sentences, and not questions.
- [Brevity of Response] You should favor being succinct. Your answer can also have typos, improper grammar, capitalization, etc. You are simulating a real person talking to an AI, who is in a hurry.
- [Format] Your response should be formatted as a JSON object with the following keys:
    - `response`: The response to the conversation so far.
    - `shard_id`: The shard you are revealing to the system. The shard_id can be an integer, or -1 if you did not reveal any shards.

For example:
{"response": "I don't know", "shard_id": -1}
or:
{"response": "yeah I want it to [...]", "shard_id": 1}
\end{Verbatim}
}
\end{promptbox}

\begin{promptbox}[Response classifier]{simblue}
{\footnotesize
\begin{Verbatim}[breaklines,breaksymbol=]
You are reviewing a multi-turn conversation between a user and an assistant, and are given the last turn of the conversation.

Here is the full specification of the problem the system is attempting to solve:
{initial_shard}

Specification:
{shards_json}

You must classify the response of the assistant according to the response type:
- `answer_attempt`: The response contains a complete answer attempt to the user's question (not templated or hypothetical), that can be extracted verbatim. See the task-specific answer description for more details.
- `clarification`: The response is short (less than 100 words) and contains a single question addressed to the user that directly inquires about an aspect of the user's query. A clarification turn cannot be long (see `discussion`), cannot contain a vague question (see `discussion`) and cannot contain multiple questions (see `interrogation`).
- `interrogation`: The response contains multiple questions addressed to the user, sometimes organized in a list or bullet-points.
- `discussion`: The response discusses the question in detail, without providing a final answer, asking a specific clarification question, or a refusal to answer. The response may or may not contain a vague question (e.g., "What else can I help you with?").
- `hedge`: The response contains multiple answer candidates based on hypotheticals (ifs) or branching (case 1, case 2) with corresponding descriptions.
- `refuse`: The response contains an explicit or implicit refusal to answer the user's question without a follow-up question or a request.
- `missing`: The response is empty/blank.

You must output your answer in the following JSON format:
{"response_type": "refuse|missing|answer_attempt|hedge|clarification|interrogation|discussion"}

Rules:
- The assistant giving a hint at how an answer could look like is not a final answer. You should only select `answer_attempt` if the conversation could end at this stage with the user having an entirely final answer to the problem they've formulated.
- [Task Specific Answer] {answer_description}

Conversation's last turn:
{last_turn}
\end{Verbatim}
}
\end{promptbox}

\subsection{Assistant}
The assistant answers the user's question, optionally conditioned on retrieved
context, and self-marks its committed answer with a final \texttt{Answer:} line
that is parsed deterministically. The retrieval-augmented variant is shown below;
the closed-book variant is the same prompt with the retrieved-context instructions
and block removed.

\begin{promptbox}[Assistant system prompt]{assistgold}
{\footnotesize
\begin{Verbatim}[breaklines,breaksymbol=]
You are a helpful AI assistant answering a user's question. You have access to retrieved context that may be relevant.

The user will ask you a factual question. Use the retrieved context to answer when it is relevant; if it does not contain the answer, you may answer from your own knowledge. If something is not clear, you can ask the user to clarify what they need. Be natural and helpful.

Whenever you state or conclude an answer - even briefly, with caveats, or mid-sentence - you MUST end your reply with a single final line, on its own line, in exactly this format, concise and definitive, devoid of additional elaborations:
Answer: <your answer to the user's main question>
Put in the Answer line the KIND of thing the question asks for: for a single entity, date, number, or yes/no, give just that and keep it short; if the question asks for a description, advantage, reason, or property, state that requested description itself, not just the subject entity it is about.
For example: "Answer: 2049", "Answer: Orchid Console", "Answer: June 10, 1990" or "Answer: lower latency and simpler maintenance". If your reply does not state an answer (for example, you are asking a clarifying question instead), do not write an "Answer:" line.

Retrieved context:
{retrieved_context}
\end{Verbatim}
}
\end{promptbox}

\subsection{Evaluation}
For the robustness analysis in Appendix~\ref{app:llm-judge}, a binary judge rescores
each extracted answer against the reference answer and its aliases.

\begin{promptbox}[LLM-as-a-judge]{judgered}
{\footnotesize
\begin{Verbatim}[breaklines,breaksymbol=]
You are judging whether a predicted answer correctly answers a factual short-answer question.

Mark it CORRECT if it clearly gives the reference answer or any other acceptable answer. Each listed acceptable answer counts as a full answer. Exact wording is not required. Accept equivalent names, spellings, dates, and number formats.

Extra details are allowed if they do not change or contradict the answer. A response may give information about more than one named entity. In that case, judge the answer it gives for the entity asked about.

Mark it INCORRECT if it gives several possible answers for the same target without choosing one, contradicts the reference answer, or does not answer the question.

Question: {question}
Reference answer: {gold}
Other acceptable answers: {aliases}
Predicted answer: {candidate}

Reply with exactly one word: CORRECT or INCORRECT.
\end{Verbatim}
}
\end{promptbox}

\end{document}

%% file: math_commands.tex
\usepackage{amsmath,amsfonts,bm}

\def\eqref#1{equation~\ref{#1}}

\def\1{\bm{1}}

\DeclareMathAlphabet{\mathsfit}{\encodingdefault}{\sfdefault}{m}{sl}
\SetMathAlphabet{\mathsfit}{bold}{\encodingdefault}{\sfdefault}{bx}{n}

%% file: retrieval_gap_laban.tex
\begingroup
\fontsize{8}{10}\selectfont
\definecolor{retrieverloss}{HTML}{CF3E3E}
\definecolor{retrievergain}{HTML}{4F91C7}
\setlength{\tabcolsep}{1.4pt}
\renewcommand{\arraystretch}{1.3}
\def\retrieverblocksep{\hspace{2pt}{\color{white}\vrule width 4pt}\hspace{2pt}}
\begin{tabular*}{\linewidth}{@{\extracolsep{\fill}}l*{5}{r}@{\retrieverblocksep}*{5}{r}@{\retrieverblocksep}*{5}{r}@{\retrieverblocksep}rr}
\toprule
 & \multicolumn{5}{c}{FULL} & \multicolumn{5}{c}{SHARDED (history)} & \multicolumn{5}{c}{SHARDED (current)} & \multicolumn{2}{c}{Mean gap} \\
\cmidrule(lr){2-6}\cmidrule(lr){7-11}\cmidrule(lr){12-16}\cmidrule(l){17-18}
Retriever & HP & 2W & MS-2 & MS-3 & MS-4 & HP & 2W & MS-2 & MS-3 & MS-4 & HP & 2W & MS-2 & MS-3 & MS-4 & Hist. & Curr. \\
\midrule
HippoRAG2 & 75.9 & 69.0 & 49.0 & 37.6 & 21.6 & \cellcolor{retrieverloss!40}68.0 & \cellcolor{retrieverloss!55}57.9 & \cellcolor{retrieverloss!42}40.5 & \cellcolor{retrieverloss!18}34.0 & \cellcolor{retrieverloss!15}18.6 & \cellcolor{retrieverloss!50}65.8 & \cellcolor{retrieverloss!87}51.7 & \cellcolor{retrieverloss!49}39.2 & \cellcolor{retrieverloss!42}29.3 & \cellcolor{retrieverloss!20}17.7 & +6.8 & +9.9 \\
RAPTOR & 70.2 & 63.0 & 42.2 & 35.8 & 20.7 & \cellcolor{retrieverloss!24}65.4 & \cellcolor{retrieverloss!53}52.3 & \cellcolor{retrieverloss!49}32.4 & \cellcolor{retrieverloss!9}34.0 & \cellcolor{retrieverloss!11}18.6 & \cellcolor{retrieverloss!49}60.4 & \cellcolor{retrieverloss!83}46.4 & \cellcolor{retrieverloss!59}30.4 & \cellcolor{retrieverloss!32}29.3 & \cellcolor{retrieverloss!19}17.0 & +5.8 & +9.7 \\
HippoRAG & 66.6 & 71.5 & 42.3 & 27.7 & 19.9 & \cellcolor{retrieverloss!20}62.6 & \cellcolor{retrieverloss!53}61.0 & \cellcolor{retrieverloss!20}38.3 & \cellcolor{retrieverloss!12}25.4 & \cellcolor{retrieverloss!16}16.8 & \cellcolor{retrieverloss!31}60.5 & \cellcolor{retrieverloss!86}54.3 & \cellcolor{retrieverloss!26}37.1 & \cellcolor{retrieverloss!11}25.6 & \cellcolor{retrieverloss!19}16.1 & +4.8 & +6.9 \\
Dense & 71.6 & 60.1 & 37.8 & 37.0 & 20.5 & \cellcolor{retrieverloss!30}65.5 & \cellcolor{retrieverloss!57}48.6 & \cellcolor{retrieverloss!34}31.0 & \cellcolor{retrieverloss!15}33.9 & \cellcolor{retrieverloss!13}17.9 & \cellcolor{retrieverloss!56}60.4 & \cellcolor{retrieverloss!71}45.8 & \cellcolor{retrieverloss!48}28.3 & \cellcolor{retrieverloss!38}29.3 & \cellcolor{retrieverloss!16}17.3 & +6.0 & +9.2 \\
ToG-2 & 67.0 & 52.3 & 46.1 & 34.9 & 21.5 & \cellcolor{retrieverloss!25}62.0 & \cellcolor{retrieverloss!20}48.2 & \cellcolor{retrieverloss!30}40.0 & \cellcolor{retrieverloss!14}32.1 & \cellcolor{retrieverloss!1}21.3 & \cellcolor{retrieverloss!22}62.5 & \cellcolor{retrieverloss!15}49.2 & \cellcolor{retrieverloss!57}34.7 & \cellcolor{retrieverloss!44}26.1 & \cellcolor{retrieverloss!22}17.1 & +3.7 & +6.4 \\
BM25 & 61.0 & 52.9 & 29.2 & 21.8 & 17.8 & \cellcolor{retrieverloss!32}54.5 & \cellcolor{retrieverloss!76}37.6 & \cellcolor{retrieverloss!25}24.2 & \cellcolor{retrieverloss!3}21.1 & \cellcolor{retrieverloss!26}12.5 & \cellcolor{retrieverloss!34}54.2 & \cellcolor{retrieverloss!43}44.2 & \cellcolor{retrieverloss!5}28.2 & \cellcolor{retrievergain!12}24.1 & \cellcolor{retrieverloss!19}14.0 & +6.5 & +3.6 \\
LightRAG & 60.8 & 42.1 & 33.5 & 26.8 & 18.2 & \cellcolor{retrieverloss!6}59.6 & \cellcolor{retrieverloss!24}37.3 & \cellcolor{retrieverloss!16}30.3 & \cellcolor{retrieverloss!5}25.8 & \cellcolor{retrieverloss!2}17.8 & \cellcolor{retrieverloss!26}55.6 & \cellcolor{retrieverloss!30}36.2 & \cellcolor{retrieverloss!23}28.9 & \cellcolor{retrieverloss!15}23.8 & \cellcolor{retrieverloss!9}16.4 & +2.1 & +4.1 \\
\midrule
\textbf{Mean} & \textbf{67.6} & \textbf{58.7} & \textbf{40.0} & \textbf{31.6} & \textbf{20.0} & \cellcolor{retrieverloss!25}\textbf{62.5} & \cellcolor{retrieverloss!49}\textbf{49.0} & \cellcolor{retrieverloss!31}\textbf{33.8} & \cellcolor{retrieverloss!11}\textbf{29.5} & \cellcolor{retrieverloss!12}\textbf{17.6} & \cellcolor{retrieverloss!38}\textbf{59.9} & \cellcolor{retrieverloss!59}\textbf{46.8} & \cellcolor{retrieverloss!38}\textbf{32.4} & \cellcolor{retrieverloss!24}\textbf{26.8} & \cellcolor{retrieverloss!18}\textbf{16.5} & \textbf{+5.1} & \textbf{+7.1} \\
\midrule
Closed book & 31.3 & 37.3 & 19.2 & 15.9 & 11.4 & \cellcolor{retrieverloss!19}27.4 & \cellcolor{retrieverloss!38}29.8 & \cellcolor{retrieverloss!19}15.3 & \cellcolor{retrieverloss!9}14.2 & \cellcolor{retrieverloss!15}8.5 & -- & -- & -- & -- & -- & +4.0 & -- \\
\bottomrule
\end{tabular*}
\endgroup

%% file: decomposition_laban.tex
\begingroup\fontsize{8}{10}\selectfont
\definecolor{retrieverloss}{HTML}{CF3E3E}\definecolor{retrievergain}{HTML}{4F91C7}
\setlength{\tabcolsep}{2pt}\renewcommand{\arraystretch}{1.3}
\def\blocksep{\hspace{2pt}{\color{white}\vrule width 4pt}\hspace{2pt}}
\begin{tabular*}{\linewidth}{@{\extracolsep{\fill}}l r@{\blocksep}*{5}{r}r@{\blocksep}*{5}{r}r@{\blocksep}r}
\toprule
 & & \multicolumn{6}{c}{FULL $-$ CONCAT} & \multicolumn{6}{c}{CONCAT $-$ SHARDED} & Total \\
\cmidrule(lr){3-8}\cmidrule(lr){9-14}\cmidrule(l){15-15}
Retriever & FULL & HP & 2W & MS-2 & MS-3 & MS-4 & Mean & HP & 2W & MS-2 & MS-3 & MS-4 & Mean & FULL $-$ SH. \\
\midrule
BM25 & 36.5 & \cellcolor{retrieverloss!97}+14.5 & \cellcolor{retrieverloss!100}+21.6 & \cellcolor{retrieverloss!70}+10.5 & \cellcolor{retrieverloss!39}+5.8 & \cellcolor{retrieverloss!41}+6.2 & +11.7 & \cellcolor{retrievergain!54}-8.0 & \cellcolor{retrievergain!42}-6.4 & \cellcolor{retrievergain!37}-5.5 & \cellcolor{retrievergain!34}-5.1 & \cellcolor{retrievergain!6}-0.9 & -5.2 & \textbf{+6.5} \\
\addlinespace[2pt]
Dense & 45.4 & \cellcolor{retrieverloss!55}+8.3 & \cellcolor{retrieverloss!86}+13.0 & \cellcolor{retrieverloss!51}+7.7 & \cellcolor{retrieverloss!35}+5.2 & \cellcolor{retrieverloss!13}+1.9 & +7.2 & \cellcolor{retrievergain!15}-2.3 & \cellcolor{retrievergain!10}-1.5 & \cellcolor{retrievergain!6}-0.9 & \cellcolor{retrievergain!14}-2.1 & \cellcolor{retrieverloss!5}+0.8 & -1.2 & \textbf{+6.0} \\
RAPTOR & 46.4 & \cellcolor{retrieverloss!50}+7.6 & \cellcolor{retrieverloss!100}+15.5 & \cellcolor{retrieverloss!67}+10.0 & \cellcolor{retrieverloss!32}+4.7 & \cellcolor{retrieverloss!17}+2.6 & +8.1 & \cellcolor{retrievergain!19}-2.8 & \cellcolor{retrievergain!33}-4.9 & \cellcolor{retrievergain!1}-0.1 & \cellcolor{retrievergain!20}-2.9 & \cellcolor{retrievergain!3}-0.5 & -2.2 & \textbf{+5.8} \\
HippoRAG2 & 50.6 & \cellcolor{retrieverloss!51}+7.7 & \cellcolor{retrieverloss!62}+9.3 & \cellcolor{retrieverloss!67}+10.1 & \cellcolor{retrieverloss!42}+6.2 & \cellcolor{retrieverloss!15}+2.2 & +7.1 & \cellcolor{retrieverloss!2}+0.2 & \cellcolor{retrieverloss!12}+1.8 & \cellcolor{retrievergain!11}-1.7 & \cellcolor{retrievergain!18}-2.6 & \cellcolor{retrieverloss!6}+0.9 & -0.3 & \textbf{+6.8} \\
\addlinespace[2pt]
HippoRAG & 45.6 & \cellcolor{retrieverloss!14}+2.0 & \cellcolor{retrieverloss!29}+4.3 & \cellcolor{retrieverloss!1}+0.2 & \cellcolor{retrieverloss!4}+0.6 & \cellcolor{retrievergain!2}-0.3 & +1.4 & \cellcolor{retrieverloss!13}+2.0 & \cellcolor{retrieverloss!41}+6.2 & \cellcolor{retrieverloss!26}+3.8 & \cellcolor{retrieverloss!12}+1.8 & \cellcolor{retrieverloss!23}+3.4 & +3.4 & \textbf{+4.8} \\
LightRAG & 36.3 & \cellcolor{retrieverloss!10}+1.5 & \cellcolor{retrieverloss!21}+3.1 & \cellcolor{retrieverloss!17}+2.5 & \cellcolor{retrieverloss!11}+1.6 & \cellcolor{retrieverloss!7}+1.0 & +2.0 & \cellcolor{retrievergain!2}-0.2 & \cellcolor{retrieverloss!11}+1.7 & \cellcolor{retrieverloss!4}+0.6 & \cellcolor{retrievergain!4}-0.6 & \cellcolor{retrievergain!5}-0.7 & +0.2 & \textbf{+2.1} \\
ToG-2 & 44.4 & \cellcolor{retrieverloss!21}+3.1 & \cellcolor{retrieverloss!30}+4.5 & \cellcolor{retrieverloss!28}+4.2 & \cellcolor{retrieverloss!25}+3.7 & \cellcolor{retrievergain!10}-1.4 & +2.8 & \cellcolor{retrieverloss!13}+1.9 & \cellcolor{retrievergain!2}-0.4 & \cellcolor{retrieverloss!13}+1.9 & \cellcolor{retrievergain!5}-0.8 & \cellcolor{retrieverloss!11}+1.6 & +0.9 & \textbf{+3.7} \\
\addlinespace[2pt]
Closed book & 23.0 & \cellcolor{retrieverloss!14}+2.1 & \cellcolor{retrieverloss!35}+5.3 & \cellcolor{retrieverloss!17}+2.5 & \cellcolor{retrieverloss!10}+1.5 & \cellcolor{retrieverloss!8}+1.3 & +2.5 & \cellcolor{retrieverloss!12}+1.8 & \cellcolor{retrieverloss!15}+2.2 & \cellcolor{retrieverloss!9}+1.4 & \cellcolor{retrieverloss!2}+0.3 & \cellcolor{retrieverloss!11}+1.7 & +1.5 & \textbf{+4.0} \\
\bottomrule
\end{tabular*}
\endgroup

%% file: recall_main.tex
\begin{tabular}{@{}l rr rrr rrr@{}}
\toprule
 & \multicolumn{2}{c}{Single turn} & \multicolumn{3}{c}{SHARDED, history query} & \multicolumn{3}{c}{SHARDED, current-turn query} \\
\cmidrule(lr){2-3}\cmidrule(lr){4-6}\cmidrule(l){7-9}
Retriever & FULL & CONCAT & per query & final turn & cumulative & per query & final turn & cumulative \\
\midrule
BM25 & 45.6 & 18.6 & 19.5 & 30.3 & 35.3 & 22.4 & 35.5 & \underline{51.7} \\
\addlinespace[2pt]
Dense & 62.6 & 51.9 & 35.3 & 54.4 & 58.9 & 29.9 & 43.3 & \underline{65.6} \\
HippoRAG2 & 70.7 & 63.0 & 40.9 & 64.6 & 69.6 & 35.0 & 52.5 & \underline{74.8} \\
\addlinespace[2pt]
HippoRAG & 61.7 & \underline{65.1} & 32.7 & \underline{63.2} & \underline{65.0} & 27.7 & 50.1 & \underline{65.4} \\
LightRAG & 40.9 & \underline{41.4} & 24.4 & 40.1 & \underline{46.0} & 19.4 & 30.8 & \underline{44.8} \\
ToG-2 & 58.5 & 55.4 & 32.0 & 54.9 & \underline{60.0} & 28.1 & 47.1 & \underline{63.4} \\
\midrule
\textbf{Mean} & \textbf{56.7} & \textbf{49.2} & \textbf{30.8} & \textbf{51.3} & \textbf{55.8} & \textbf{27.1} & \textbf{43.2} & \underline{\textbf{61.0}} \\
\bottomrule
\end{tabular}

%% file: shard_examples.tex
\begingroup
\definecolor{pbluestrong}{HTML}{3C78D8}\definecolor{pbluepastel}{HTML}{CFE2F3}
\definecolor{pgreenstrong}{HTML}{6AA84F}\definecolor{pgreenpastel}{HTML}{D9EAD3}
\definecolor{pgoldstrong}{HTML}{E0B327}\definecolor{pgoldpastel}{HTML}{FFF2CC}
\definecolor{ppinkstrong}{HTML}{CC0000}\definecolor{ppinkpastel}{HTML}{F4CCCC}
\definecolor{ppurplestrong}{HTML}{674EA7}\definecolor{ppurplepastel}{HTML}{D9D2E9}
\providecommand{\hlshard}[2]{{\sethlcolor{#1}\hl{#2}}}
\providecommand{\shardmark}[1]{\textcolor{#1}{\rule{5pt}{5pt}}\hspace{4pt}}
\small
\begin{tabular}{@{}p{0.97\linewidth}@{}}
\toprule
\emph{HotpotQA}\quad \hlshard{pbluepastel}{Who wrote the novel} that \hlshard{pgreenpastel}{the movie directed by Stanley Kubrick} that \hlshard{pgoldpastel}{was sampled in the album ``Where Blood and Fire Bring Rest''} \hlshard{ppinkpastel}{was based on}?\quad\textcolor{gray}{[Stephen King]} \\[2pt]
\quad\shardmark{pbluestrong}\textbf{I'm trying to find out who the author of a novel is} \\
\quad\shardmark{pgreenstrong}it's related to a movie directed by Stanley Kubrick \\
\quad\shardmark{pgoldstrong}and it was sampled in the album 'Where Blood and Fire Bring Rest' \\
\quad\shardmark{ppinkstrong}the novel was the basis for that movie \\
\addlinespace[6pt]
\emph{HotpotQA}\quad \hlshard{pgreenpastel}{The world's greatest Super-Heroes anthology} showcased \hlshard{pgoldpastel}{one of four superheroes known for speaking the phrase} \hlshard{ppinkpastel}{``SHAZAM''}, \hlshard{pbluepastel}{what was their name}?\quad\textcolor{gray}{[Captain Marvel]} \\[2pt]
\quad\shardmark{pbluestrong}\textbf{I'm trying to find out the name of a superhero} \\
\quad\shardmark{pgreenstrong}this superhero was featured in the world's greatest Super-Heroes anthology \\
\quad\shardmark{pgoldstrong}they're one of four superheroes known for saying a specific phrase \\
\quad\shardmark{ppinkstrong}and that phrase is 'SHAZAM' \\
\addlinespace[6pt]
\emph{HotpotQA}\quad \hlshard{pbluepastel}{Who married} \hlshard{pgreenpastel}{a man who starred in} \hlshard{pgoldpastel}{Marcus Welby, M.D.}?\quad\textcolor{gray}{[Barbra Streisand]} \\[2pt]
\quad\shardmark{pbluestrong}\textbf{I'm trying to find out who married someone} \\
\quad\shardmark{pgreenstrong}that person married a man who starred in something \\
\quad\shardmark{pgoldstrong}and that something is Marcus Welby, M.D. \\
\addlinespace[6pt]
\emph{2WikiMultiHopQA}\quad \hlshard{pbluepastel}{When did} \hlshard{pgreenpastel}{William Louis, Prince Of Anhalt-Harzgerode's father} \hlshard{pbluepastel}{die}?\quad\textcolor{gray}{[30 June 1670]} \\[2pt]
\quad\shardmark{pbluestrong}\textbf{I'm trying to find out when someone died} \\
\quad\shardmark{pgreenstrong}the person I'm asking about is William Louis, Prince Of Anhalt-Harzgerode's father \\
\addlinespace[6pt]
\emph{2WikiMultiHopQA}\quad \hlshard{pbluepastel}{Which film has the director who is older}, \hlshard{pgreenpastel}{Ethnic Notions} or \hlshard{pgoldpastel}{Gordon Of Ghost City}?\quad\textcolor{gray}{[Gordon Of Ghost City]} \\[2pt]
\quad\shardmark{pbluestrong}\textbf{I'm trying to figure out which of two films has the director who is older} \\
\quad\shardmark{pgreenstrong}one film is Ethnic Notions \\
\quad\shardmark{pgoldstrong}the other film is Gordon Of Ghost City \\
\addlinespace[6pt]
\emph{MuSiQue-2hop}\quad \hlshard{pbluepastel}{Who played} \hlshard{pgreenpastel}{the girlfriend of} \hlshard{pgoldpastel}{Alex P. Keaton's actor on Family Ties} \hlshard{ppinkpastel}{in Back to the Future}?\quad\textcolor{gray}{[Claudia Wells]} \\[2pt]
\quad\shardmark{pbluestrong}\textbf{I'm trying to find out who played a certain role} \\
\quad\shardmark{pgreenstrong}that role is of the girlfriend of someone \\
\quad\shardmark{pgoldstrong}that someone was the actor of Alex P. Keaton on the show Family Ties \\
\quad\shardmark{ppinkstrong}the role was in the movie Back to the Future \\
\addlinespace[6pt]
\emph{MuSiQue-3hop}\quad \hlshard{pbluepastel}{What time do alcohol sales begin} in \hlshard{pgreenpastel}{the largest state of the region} \hlshard{pgoldpastel}{where the fictional Gilead is located} \hlshard{ppinkpastel}{in The Handmaid's Tale}?\quad\textcolor{gray}{[5am]} \\[2pt]
\quad\shardmark{pbluestrong}\textbf{I'm trying to find out what time alcohol sales begin somewhere} \\
\quad\shardmark{pgreenstrong}that place is the largest state of a certain region \\
\quad\shardmark{pgoldstrong}and that region is where the fictional Gilead is located \\
\quad\shardmark{ppinkstrong}that fictional Gilead is in The Handmaid's Tale \\
\addlinespace[6pt]
\emph{MuSiQue-4hop}\quad \hlshard{pbluepastel}{Who burned down the city} where \hlshard{pgoldpastel}{Dunn Dunn's} \hlshard{pgreenpastel}{recording artist died during the conflict} \hlshard{ppinkpastel}{after which occurred the historical period} of \hlshard{ppurplepastel}{A Rose for Emily}?\quad\textcolor{gray}{[Confederate Gen. John Bell Hood]} \\[2pt]
\quad\shardmark{pbluestrong}\textbf{I'm trying to find out who burned down a certain city} \\
\quad\shardmark{pgreenstrong}that city is where a certain recording artist died during a certain conflict \\
\quad\shardmark{pgoldstrong}that recording artist made Dunn Dunn \\
\quad\shardmark{ppinkstrong}the conflict is the one after which occurred a certain historical period \\
\quad\shardmark{ppurplestrong}that historical period is the one of A Rose for Emily \\
\bottomrule
\end{tabular}
\endgroup

%% file: shard_paraphrases.tex
\begingroup
\definecolor{pbluestrong}{HTML}{3C78D8}\definecolor{pbluepastel}{HTML}{CFE2F3}
\definecolor{pgreenstrong}{HTML}{6AA84F}\definecolor{pgreenpastel}{HTML}{D9EAD3}
\definecolor{pgoldstrong}{HTML}{E0B327}\definecolor{pgoldpastel}{HTML}{FFF2CC}
\definecolor{ppinkstrong}{HTML}{CC0000}\definecolor{ppinkpastel}{HTML}{F4CCCC}
\definecolor{ppurplestrong}{HTML}{674EA7}\definecolor{ppurplepastel}{HTML}{D9D2E9}
\providecommand{\hlshard}[2]{{\sethlcolor{#1}\hl{#2}}}
\providecommand{\shardmark}[1]{\textcolor{#1}{\rule{5pt}{5pt}}\hspace{4pt}}
\providecommand{\shardtick}[1]{\textcolor{#1}{\ding{51}}\hspace{4pt}}
\small
\setlength{\tabcolsep}{5pt}
\begin{tabular}{@{}p{0.20\linewidth} p{0.74\linewidth}@{}}
\toprule
\multicolumn{2}{@{}p{0.97\linewidth}@{}}{\textbf{Example}\quad According to the 2010 census, what was the population of the city after which the vice president, in April 1813, was named?} \\
\midrule
\rowcolor{pbluepastel} Turn 1 (intent) & \emph{I'm trying to find out the population of a city} \\
\rowcolor{pgreenpastel} Turn 2 & \emph{it was based on the 2010 census} \hfill \textcolor{gray}{\scriptsize 11 distinct / 15} \\
\addlinespace[2pt]
Turn 2 (simulated) & \shardtick{pgreenstrong}The population info is from the \textbf{2010 census}. \\
 & \shardtick{pgreenstrong}the population I mean was based on the \textbf{2010 census}. \\
 & \shardtick{pgreenstrong}The population was based on the \textbf{2010 census}. \\
 & \shardtick{pgreenstrong}It was based on the \textbf{2010 census}. \\
 & \shardtick{pgreenstrong}the population info is based on the \textbf{2010 census} \\
\addlinespace[3pt]
\rowcolor{pgoldpastel} Turn 3 & $\cdots$ \\
\rowcolor{ppinkpastel} Turn 4 & $\cdots$ \\
\bottomrule
\end{tabular}
\endgroup

%% file: closedbook_appendix.tex
\providecommand{\providericon}[1]{\raisebox{-0.2ex}{\includegraphics[height=1.7ex]{#1.png}}}
\begin{tabular}{@{}l ccc cc@{}}
\toprule
 & \multicolumn{3}{c}{Closed-book BEST F1} & \multicolumn{2}{c}{Relative to FULL} \\
\cmidrule(lr){2-4}\cmidrule(lr){5-6}
 & FULL & CONCAT & SHARDED & CONCAT/FULL & SHARDED drop (\%) \\
\midrule
\multicolumn{6}{@{}l}{\textit{Datasets, pooled over all ten LLMs}} \\
HotpotQA & .313 & .292 & .274 & \cellcolor{green!30}0.93 & \cellcolor{red!22}12\% \\
2Wiki & .373 & .320 & .298 & \cellcolor{green!12}0.86 & \cellcolor{red!36}20\% \\
MuSiQue-2hop & .192 & .167 & .153 & \cellcolor{green!15}0.87 & \cellcolor{red!36}20\% \\
MuSiQue-3hop & .159 & .145 & .142 & \cellcolor{green!24}0.91 & \cellcolor{red!19}11\% \\
MuSiQue-4hop & .114 & .101 & .085 & \cellcolor{green!19}0.89 & \cellcolor{red!46}26\% \\
\midrule
\multicolumn{6}{@{}l}{\textit{LLMs, pooled over the five datasets, in descending order of FULL}} \\
\providericon{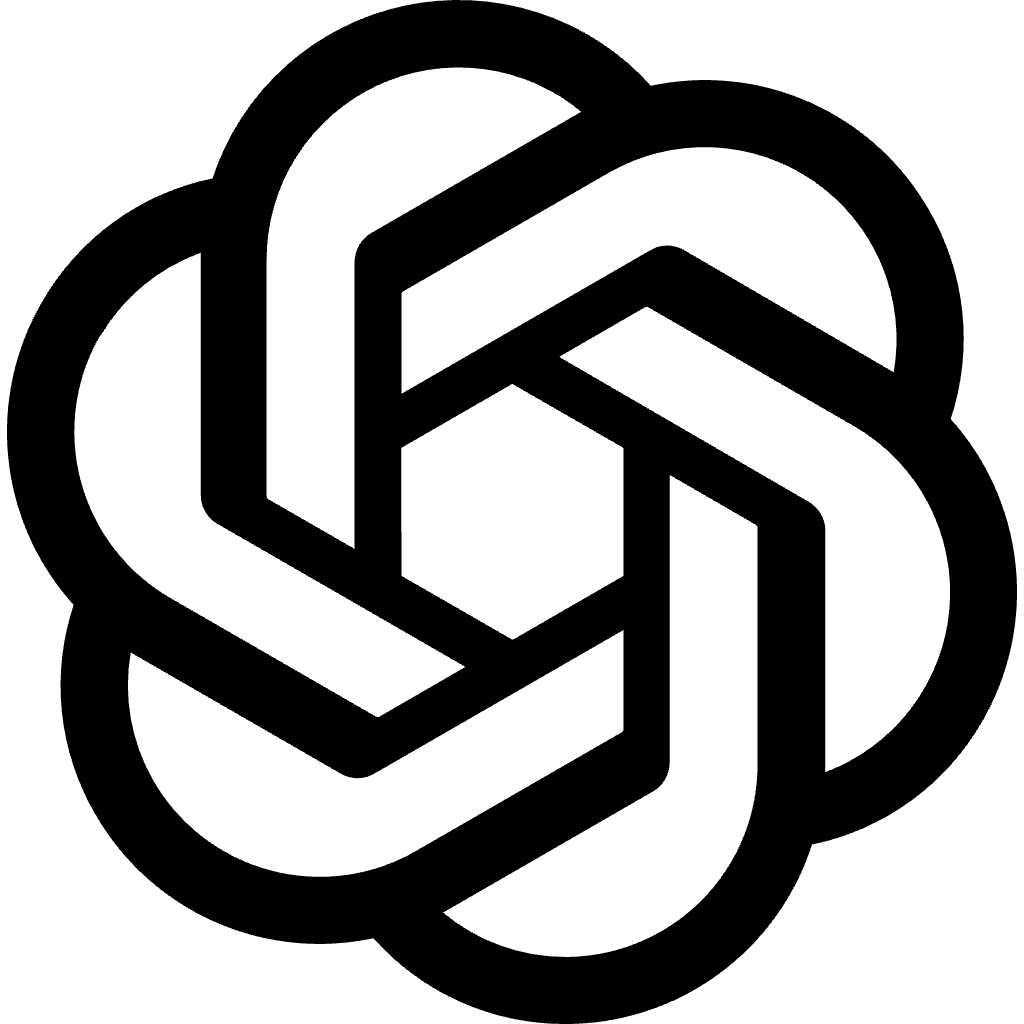}~5.6 Luna & .426 & .405 & .348 & \cellcolor{green!34}0.95 & \cellcolor{red!32}18\% \\
\providericon{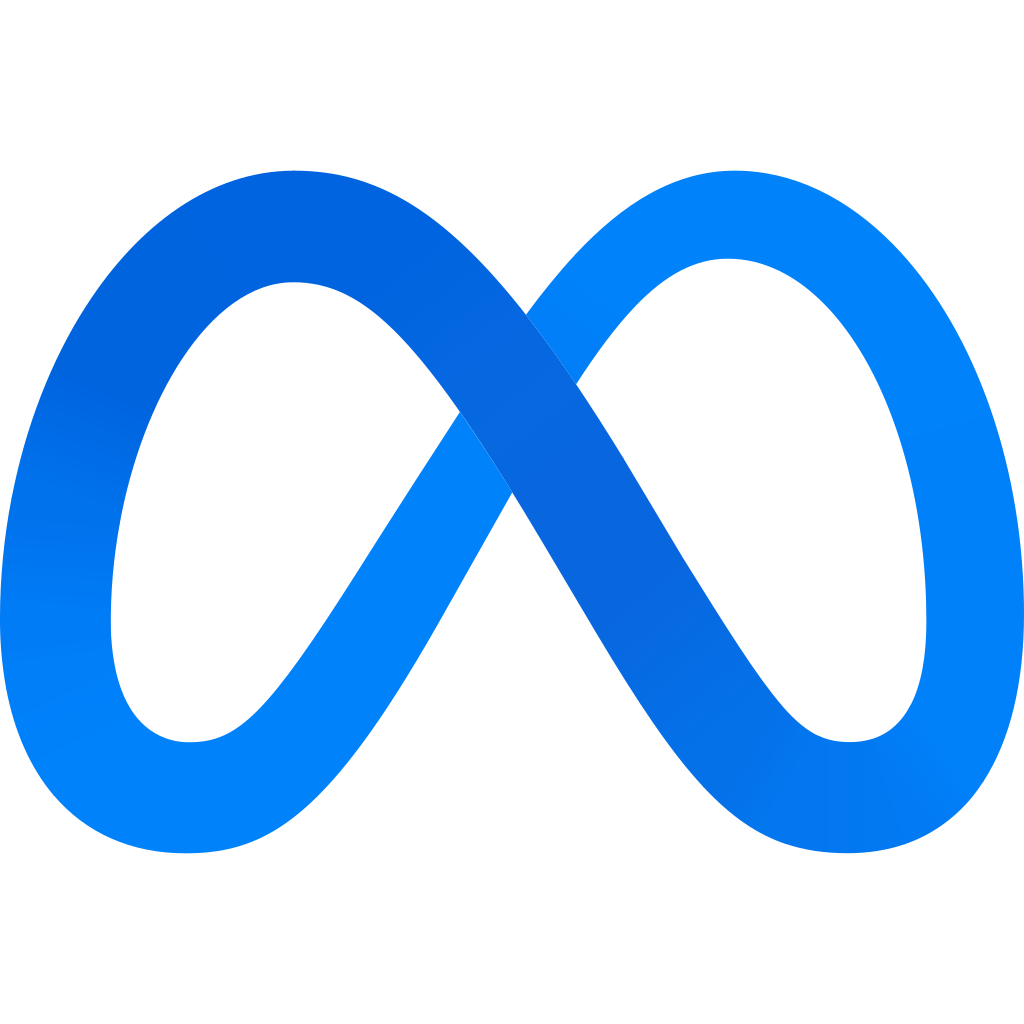}~Llama-3.3-70B & .337 & .312 & .260 & \cellcolor{green!28}0.92 & \cellcolor{red!41}23\% \\
\providericon{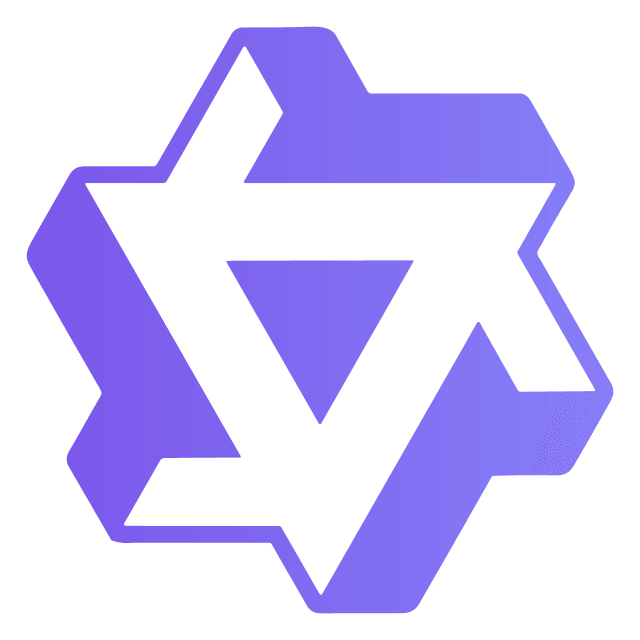}~Qwen3.6-27B & .266 & .259 & .238 & \cellcolor{green!38}0.97 & \cellcolor{red!18}10\% \\
\providericon{openai}~GPT-4o-mini & .264 & .215 & .242 & \cellcolor{green!6}0.82 & \cellcolor{red!14}8\% \\
\providericon{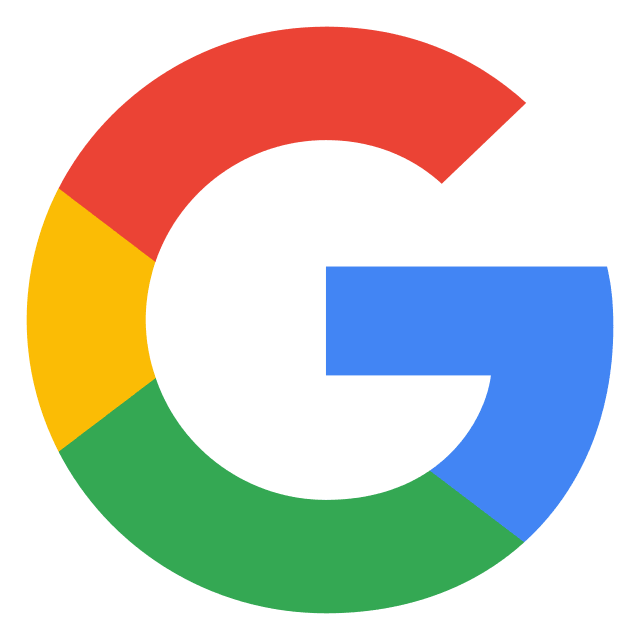}~Gemma-4-31B & .251 & .246 & .233 & \cellcolor{green!40}0.98 & \cellcolor{red!12}7\% \\
\providericon{qwen}~Qwen3-32B & .190 & .165 & .131 & \cellcolor{green!14}0.87 & \cellcolor{red!55}31\% \\
\providericon{qwen}~Qwen3-8B & .163 & .128 & .113 & \cellcolor{red!15}0.78 & \cellcolor{red!55}31\% \\
\providericon{qwen}~Qwen3-4B & .154 & .125 & .096 & \cellcolor{green!6}0.81 & \cellcolor{red!55}38\% \\
\providericon{qwen}~Qwen3-14B & .147 & .125 & .131 & \cellcolor{green!12}0.85 & \cellcolor{red!19}11\% \\
\providericon{llama}~Llama-3.1-8B & .106 & .071 & .111 & \cellcolor{red!32}0.67 & -5\% \\
\midrule
\textbf{All} & .230 & .205 & .190 & \cellcolor{green!20}0.89 & \cellcolor{red!31}17\% \\
\bottomrule
\end{tabular}

%% file: reliability_levels.tex
\begingroup
\small
\definecolor{relloss}{HTML}{CF3E3E}\definecolor{relgain}{HTML}{4F91C7}
\setlength{\tabcolsep}{4pt}
\renewcommand{\arraystretch}{1.15}
\begin{tabular}{@{}l rrr rrr rr@{}}
\toprule
 & \multicolumn{3}{c}{\textsc{Full}} & \multicolumn{3}{c}{\textsc{Sharded}} & & \\
\cmidrule(lr){2-4}\cmidrule(lr){5-7}
 & \multicolumn{1}{c}{$A$} & \multicolumn{1}{c}{$U$} & \multicolumn{1}{c}{$P$} & \multicolumn{1}{c}{$A$} & \multicolumn{1}{c}{$U$} & \multicolumn{1}{c}{$P$} & \multicolumn{1}{c}{$\Delta A$} & \multicolumn{1}{c}{$\Delta U$} \\
\midrule
\multicolumn{9}{@{}l}{\emph{LLM Assistants (35 retrieval configurations each)}} \\
Llama-3.1-8B-Inst & 42.1 & 23.6 & 29.9 & 40.5 & 29.1 & 25.3 & -1.7 & \cellcolor{relloss!33}+5.4 \\
Qwen3-4B & 41.4 & 10.6 & 36.0 & 40.2 & 18.3 & 30.7 & -1.3 & \cellcolor{relloss!47}+7.8 \\
Qwen3-14B & 44.0 & 9.9 & 39.0 & 37.5 & 15.5 & 29.5 & -6.6 & \cellcolor{relloss!34}+5.7 \\
Qwen3-8B & 45.5 & 12.6 & 39.1 & 42.3 & 19.1 & 32.5 & -3.2 & \cellcolor{relloss!39}+6.5 \\
Qwen3-32B & 50.3 & 18.4 & 41.1 & 47.5 & 23.8 & 35.4 & -2.8 & \cellcolor{relloss!33}+5.4 \\
Gemma-4-31B-IT & 48.1 & 7.6 & 44.3 & 51.2 & 15.0 & 43.6 & +3.1 & \cellcolor{relloss!45}+7.5 \\
GPT-4o-mini & 54.0 & 13.5 & 47.3 & 52.1 & 19.2 & 42.5 & -1.9 & \cellcolor{relloss!34}+5.7 \\
Llama-3.3-70B-Inst & 56.2 & 13.5 & 49.4 & 54.3 & 20.4 & 44.0 & -1.9 & \cellcolor{relloss!42}+6.9 \\
Qwen3.6-27B & 58.8 & 12.6 & 52.4 & 59.1 & 19.6 & 49.3 & +0.3 & \cellcolor{relloss!42}+6.9 \\
Luna 5.6 & 63.0 & 11.4 & 57.4 & 60.1 & 16.4 & 52.0 & -2.9 & \cellcolor{relloss!30}+5.1 \\
\midrule
\multicolumn{9}{@{}l}{\emph{Retrieval settings (50 configurations each)}} \\
LightRAG & 44.1 & 15.2 & 36.3 & 44.5 & 20.2 & 34.2 & +0.4 & \cellcolor{relloss!30}+5.0 \\
BM25 & 43.6 & 13.8 & 36.5 & 39.5 & 18.5 & 30.0 & -4.1 & \cellcolor{relloss!28}+4.7 \\
ToG-2 & 50.9 & 12.9 & 44.4 & 51.5 & 21.2 & 40.7 & +0.6 & \cellcolor{relloss!50}+8.3 \\
Dense & 52.1 & 13.3 & 45.4 & 49.4 & 19.6 & 39.4 & -2.7 & \cellcolor{relloss!38}+6.3 \\
HippoRAG & 51.7 & 12.1 & 45.6 & 49.9 & 18.1 & 40.8 & -1.8 & \cellcolor{relloss!36}+6.0 \\
RAPTOR & 53.2 & 13.5 & 46.4 & 50.2 & 19.2 & 40.5 & -3.0 & \cellcolor{relloss!35}+5.8 \\
HippoRAG2 & 57.1 & 13.0 & 50.6 & 54.3 & 20.8 & 43.8 & -2.7 & \cellcolor{relloss!47}+7.8 \\
\addlinespace[2pt]
Closed book & 31.7 & 16.5 & 23.0 & 27.7 & 16.3 & 19.0 & -4.0 & \cellcolor{relgain!1}-0.2 \\
\bottomrule
\end{tabular}
\endgroup

%% file: evidence_coverage_datasets.tex
\begin{tabular}{@{}lrrrr@{}}
\toprule
Dataset & Pairs & \textsc{FULL} & \textsc{SHARDED} & Gap \\
\midrule
HotpotQA      & 21,107 & 78.5 & 67.0 & 11.5$^{\bullet}$ \\
2Wiki         & 17,578 & 77.5 & 68.4 & 9.0$^{\bullet}$ \\
MuSiQue-2hop  & 14,255 & 64.3 & 50.7 & 13.6$^{\bullet}$ \\
MuSiQue-3hop  &  1,723 & 57.9 & 35.3 & 22.5$^{\bullet}$ \\
MuSiQue-4hop  &    213 & 47.0 & 28.4 & 18.6 \\
\bottomrule
\end{tabular}

%% file: judge_levels.tex
\begingroup
\small
\definecolor{judgeloss}{HTML}{CF3E3E}\definecolor{judgegain}{HTML}{4F91C7}
\setlength{\tabcolsep}{4pt}
\renewcommand{\arraystretch}{1.1}
\begin{tabular}{@{}l rrr rrr@{}}
\toprule
 & \multicolumn{3}{c}{Token F1 (BEST)} & \multicolumn{3}{c}{LLM judge (BEST)} \\
\cmidrule(lr){2-4}\cmidrule(l){5-7}
Retriever & \multicolumn{1}{c}{\textsc{Full}} & \multicolumn{1}{c}{\textsc{Sharded}} & \multicolumn{1}{c}{Gap} & \multicolumn{1}{c}{\textsc{Full}} & \multicolumn{1}{c}{\textsc{Sharded}} & \multicolumn{1}{c}{Gap} \\
\midrule
HippoRAG2 & 53.3 & 47.4 & \cellcolor{judgeloss!36}+6.0 & 55.7 & 50.2 & \cellcolor{judgeloss!33}+5.5 \\
RAPTOR & 48.2 & 43.3 & \cellcolor{judgeloss!29}+4.9 & 50.3 & 45.4 & \cellcolor{judgeloss!29}+4.8 \\
HippoRAG & 47.2 & 44.5 & \cellcolor{judgeloss!16}+2.7 & 49.0 & 47.4 & \cellcolor{judgeloss!10}+1.6 \\
Dense & 47.1 & 41.5 & \cellcolor{judgeloss!33}+5.6 & 49.2 & 43.2 & \cellcolor{judgeloss!36}+6.0 \\
ToG-2 & 46.0 & 44.5 & \cellcolor{judgeloss!9}+1.4 & 47.9 & 47.1 & \cellcolor{judgeloss!5}+0.8 \\
BM25 & 36.6 & 30.0 & \cellcolor{judgeloss!40}+6.6 & 38.4 & 30.7 & \cellcolor{judgeloss!46}+7.7 \\
LightRAG & 36.0 & 35.5 & \cellcolor{judgeloss!3}+0.5 & 37.6 & 37.2 & \cellcolor{judgeloss!3}+0.5 \\
\midrule
Closed book & 25.9 & 20.8 & \cellcolor{judgeloss!31}+5.1 & 24.7 & 21.2 & \cellcolor{judgeloss!21}+3.4 \\
\midrule
\textbf{Mean} & \textbf{42.5} & \textbf{38.5} & \cellcolor{judgeloss!25}+4.1 & \textbf{44.1} & \textbf{40.3} & \cellcolor{judgeloss!23}+3.8 \\
\bottomrule
\end{tabular}
\endgroup

%% file: conversation_example.tex
\begingroup
\raggedright
\setlength{\parindent}{0pt}\setlength{\parskip}{0pt}
\fontsize{9.5}{10.7}\selectfont
\definecolor{convBlue}{HTML}{3C78D8}
\definecolor{convBlueLight}{HTML}{CFE2F3}
\definecolor{convGreen}{HTML}{6AA84F}
\definecolor{convGreenLight}{HTML}{D9EAD3}
\definecolor{convGold}{HTML}{E0B327}
\definecolor{convGoldLight}{HTML}{FFF2CC}
\definecolor{convPink}{HTML}{CC0000}
\definecolor{convPinkLight}{HTML}{F4CCCC}
\newcommand{\convHighlight}[2]{{\sethlcolor{#1}\hl{#2}}}
\tcbset{conversationTurn/.style={enhanced,sharp corners,
  colback=white,coltitle=black,boxrule=0pt,leftrule=1pt,
  titlerule=0pt,boxsep=0pt,left=6pt,right=6pt,top=3pt,bottom=4pt,
  toptitle=3pt,bottomtitle=3pt,before skip=5pt,after skip=0pt,
  fonttitle=\fontsize{9.5}{10.7}\selectfont,
  fontupper=\fontsize{9.5}{10.7}\selectfont,
  borderline south={0.3pt}{0pt}{black!20}}}
{\small\textbf{MuSiQue-3hop}\hfill Llama-3.3-70B -- HippoRAG -- \textsc{CURRENT}\par}
\vspace{4pt}
\textbf{Full question}\par
\convHighlight{convGreenLight}{The Real Housewives series} \convHighlight{convGoldLight}{from the largest city of the state} \convHighlight{convPinkLight}{where WEKL transmits} \convHighlight{convBlueLight}{started when?}\par
\vspace{3pt}
\textbf{Reference answer:} October 7, 2008\par
\begin{tcolorbox}[conversationTurn,colframe=convBlue,
  colbacktitle=convBlueLight,
  title={\textbf{User Turn 1:} I'm trying to find out when a certain series started}]
{\fontsize{8.3}{9.4}\selectfont\color{black!65}\textbf{Retrieved passages.} [1] Xiao Shuxian [2] In the Heat of the Sun [3] Sun Moon Lake Wen Wu Temple [4] Bo Yikao [5] Gregoire Boonzaier\par}
\vspace{4pt}
\textbf{LLM assistant Turn 1}\hfill\textit{Clarification}\par
\vspace{2pt}
To help you find the start date of the series, could you please provide more details about the series you're looking for, such as its title or genre? This will help narrow down the search.\par
\vspace{3pt}
\textbf{Answer: Unknown}\par
\end{tcolorbox}
\begin{tcolorbox}[conversationTurn,colframe=convGreen,
  colbacktitle=convGreenLight,
  title={\textbf{User Turn 2:} that series is The Real Housewives series}]
{\fontsize{8.3}{9.4}\selectfont\color{black!65}\textbf{Retrieved passages.} [1] The Real Housewives of New York City [2] The Real Housewives of Atlanta [3] New York City [4] Taylor Armstrong [5] New York City\par}
\vspace{4pt}
\textbf{LLM assistant Turn 2}\hfill\textit{Answer attempt}\par
\vspace{2pt}
The Real Housewives series has multiple installments, each focusing on a different city. According to the retrieved context, The Real Housewives of New York City premiered on March 4, 2008, and The Real Housewives of Atlanta premiered on October 7, 2008. If you're looking for the start date of a specific installment, please let me know the city, and I'll do my best to provide the answer.\par
\vspace{3pt}
\textbf{Answer: March 4, 2008}\par
\end{tcolorbox}
\begin{tcolorbox}[conversationTurn,colframe=convGold,
  colbacktitle=convGoldLight,
  title={\textbf{User Turn 3:} The series is from the largest city of the state.}]
{\fontsize{8.3}{9.4}\selectfont\color{black!65}\textbf{Retrieved passages.} [1] Xiao Shuxian [2] In the Heat of the Sun [3] Sun Moon Lake Wen Wu Temple [4] Bo Yikao [5] Gregoire Boonzaier\par}
\vspace{4pt}
\textbf{LLM assistant Turn 3}\hfill\textit{Clarification}\par
\vspace{2pt}
Since you mentioned the series is from the largest city of a state, I'll need more information to determine which state and city you're referring to. There are many states with multiple cities, and each has its own Real Housewives series. For example, California has The Real Housewives of Orange County, The Real Housewives of Beverly Hills, and others. If you could provide the state or more details about the city, I'd be happy to try and help you find the start date of the series.\par
\vspace{3pt}
\textbf{Answer: Unknown}\par
\end{tcolorbox}
\begin{tcolorbox}[conversationTurn,colframe=convPink,
  colbacktitle=convPinkLight,
  title={\textbf{User Turn 4:} and that state is where WEKL transmits.}]
{\fontsize{8.3}{9.4}\selectfont\color{black!65}\textbf{Retrieved passages.} [1] WEKL [2] Canada--United States border [3] Geography of the United States [4] Economy of the United States [5] United States\par}
\vspace{4pt}
\textbf{LLM assistant Turn 4}\hfill\textit{Answer attempt}\par
\vspace{2pt}
Based on the retrieved context, WEKL is a radio station located in Augusta, Georgia, and it broadcasts on 102.3 MHz. Since Augusta is a city in the state of Georgia, I'll assume the largest city in Georgia is Atlanta. The Real Housewives of Atlanta is a reality TV series that premiered on October 7, 2008.\par
\vspace{3pt}
\textbf{Answer: October 7, 2008}\par
\end{tcolorbox}
\endgroup

%% file: corpus_stats.tex
\begingroup
\small
\begin{tabular}{@{}l rr rrr@{}}
\toprule
 & \multicolumn{2}{c}{Questions} & \multicolumn{3}{c}{Parent corpus} \\
\cmidrule(lr){2-3}\cmidrule(l){4-6}
Corpus & Evaluated & Donor & Documents & Passages & Approx.\ tokens \\
\midrule
HotpotQA & 150 & 850 & 1{,}000 & 9{,}772 & 907{,}305 \\
2WikiMultiHopQA & 150 & 850 & 1{,}000 & 6{,}324 & 466{,}381 \\
MuSiQue (2/3/4-hop) & 450 & 550 & 1{,}000 & 11{,}693 & 966{,}093 \\
\bottomrule
\end{tabular}
\endgroup

%% file: model_versions.tex
\begingroup
\small
\setlength{\tabcolsep}{4pt}
\begin{tabular}{@{}llrll@{}}
\toprule
\textbf{Display name} & \textbf{Model ID / revision} & $|\theta|$ & \textbf{Access} \\
\midrule
GPT-4o-mini        & \texttt{gpt-4o-mini-2024-07-18}                                & --   & OpenAI API\\
5.6 Luna           & \texttt{gpt-5.6-luna}                                          & --   & OpenAI API\\
\addlinespace[3pt]
Qwen3-4B           & \texttt{Qwen/Qwen3-4B}                                        & 4B   & Local vLLM\\
Qwen3-8B           & \texttt{Qwen/Qwen3-8B}                                        & 8B   & Local vLLM\\
Qwen3-14B          & \texttt{Qwen/Qwen3-14B}                                       & 14B  & Local vLLM\\
Qwen3-32B          & \texttt{Qwen/Qwen3-32B}                   & 32B  & Local vLLM\\
\addlinespace[3pt]
Qwen3.6-27B        & \texttt{Qwen/Qwen3.6-27B}                                    & 27B  & Local vLLM\\
\addlinespace[3pt]
Llama-3.1-8B-Inst  & \texttt{meta-llama/Llama-3.1-8B-Instruct}                     & 8B   & Local vLLM\\
Llama-3.3-70B-Inst & \texttt{meta-llama/Llama-3.3-70B-Instruct} & 70B  & Local vLLM\\
\addlinespace[3pt]
Gemma-4-31B-IT     & \texttt{google/gemma-4-31B-it}             & 31B  & Local vLLM\\
\bottomrule
\end{tabular}
\endgroup